%% file: halogs.tex
\documentclass{article}

\usepackage[preprint]{style}

\usepackage[utf8]{inputenc} 
\usepackage[T1]{fontenc}    
\usepackage{hyperref}       
\usepackage{url}            
\usepackage{booktabs}       
\usepackage{amsfonts}       
\usepackage{nicefrac}       
\usepackage{microtype}      
\usepackage{xcolor}         
\usepackage{graphicx} 
\usepackage{threeparttable}
\usepackage{multirow}
\usepackage{mathtools}
\usepackage{tabularx}
\usepackage{color}
\usepackage{colortbl}
\usepackage{amsthm}
\usepackage{caption}
\usepackage{marvosym}
\usepackage{enumitem}
\usepackage{amssymb}
\usepackage{bm}

\usepackage{gensymb}
\usepackage{makecell}
\usepackage{svg} 
\usepackage{float}
\usepackage{indentfirst}
 
\usepackage{multicol}
\usepackage{xspace}

\usepackage{amsmath}
\usepackage{overpic}
\usepackage{wrapfig}
\usepackage{array}

\newcommand{\modelname}{HaloGS\xspace}

\title{HaloGS: Loose Coupling of Compact Geometry and Gaussian Splats for 3D Scenes}

\author{Changjian Jiang $^{1,2}$\thanks{Equal contribution} \quad Kerui Ren$^{2,3}$\protect\footnotemark[1] \quad Linning Xu$^4$ \quad Jiong Chen$^{5}$\\ \textbf{Jiangmiao Pang$^2$ \quad Yu Zhang$^1$\thanks{Corresponding author}
\quad Bo Dai$^{6}$ \quad Mulin Yu$^{2}$}\\
$^1$Zhejiang University, $^2$Shanghai Artificial Intelligence Laboratory, \\$^3$Shanghai Jiao Tong University, $^4$The Chinese University of Hong Kong, \\$^5$Inria, $^6$ The University of Hong Kong
}

\begin{document}

\maketitle
{
\begin{center}
    \vspace{-2em}
    \centering
    \captionsetup{type=figure}
    \includegraphics[width=1.0\textwidth]{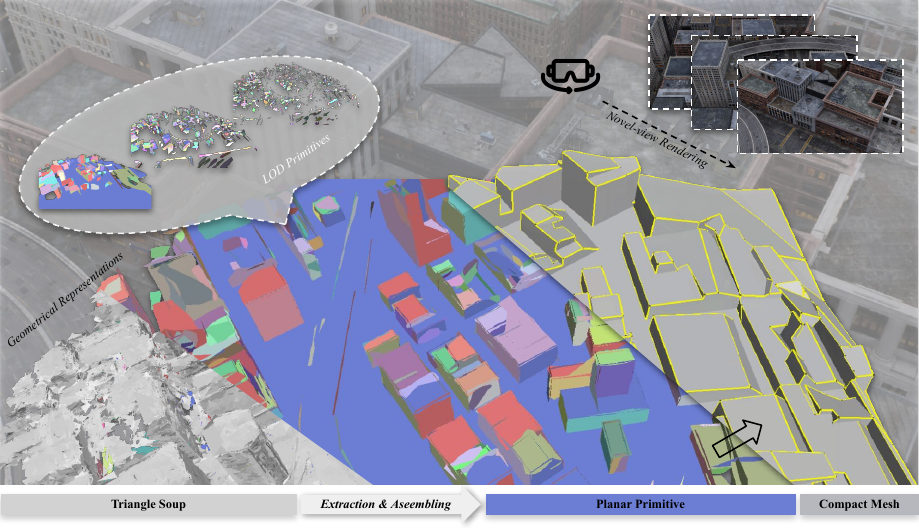}
    \captionof{figure}
    {
    \modelname presents a dual‐representation framework that disentangles geometry from appearance for multiview reconstruction. 
    Geometrically, it represents scene structure as a surface‐aligned triangle soup and augments it with neural Gaussians for photorealistic rendering. 
    From this soup, we progressively extract planar primitives and assemble them into compact meshes.
    HaloGS combines high‑fidelity appearance with lightweight geometry for efficient storage and downstream processing.
    The flexible triangle primitive makes it handle both indoor and outdoor environments easily, adapting seamlessly to varying levels of detail and scene complexity.  
    Here, we illustrate with the large-scale MatrixCity~\cite{li2023matrixcity} scene, please visit our project page for additional results: \url{https://city-super.github.io/halogs/} . 
    }
    \label{fig:reaser}
\end{center}
}
   
\input{sec/abstract}
\input{sec/introduction}
\input{sec/related_work}
\input{sec/method}
\input{sec/experiments}
\input{sec/conclusion}

\bibliographystyle{plain}
\bibliography{ref}


\appendix
\input{sec/supplementary}


\end{document}

%% file: sec/abstract.tex
\begin{abstract}
High‑fidelity 3D reconstruction and rendering hinge on capturing precise geometry while preserving photo‑realistic detail.
Most existing methods either fuse these goals into a single cumbersome model or adopt hybrid schemes whose uniform primitives lead to a trade‑off between efficiency and fidelity.
In this paper, we introduce HaloGS, a dual‑representation that loosely couples coarse triangles for geometry with Gaussian primitives for appearance, motivated by the lightweight classic geometry representations and their proven efficiency in real‑world applications.
Our design yields a compact yet expressive model capable of photo‑realistic rendering across both indoor and outdoor environments, seamlessly adapting to varying levels of scene complexity. 
Experiments on multiple benchmark datasets demonstrate that our method yields both compact, accurate geometry and high-fidelity renderings, especially in challenging scenarios where robust geometric structure make a clear difference.
\end{abstract}

%% file: sec/introduction.tex
\section{Introduction}
\label{sec:intro}

Photorealistic 3D scene rendering has been revolutionized by neural representations such as Neural Radiance Fields (NeRF) \cite{mildenhall2021nerf}. 
Recently, 3D Gaussian Splatting (3DGS) \cite{kerbl20233d} introduced an explicit point-based scene model that combines the strengths of traditional primitives with neural rendering, attaining state-of-the-art image quality and real-time performance. However, 3DGS models often rely on an excessive number of Gaussian primitives to fit every training view, resulting in redundant elements and a lack of coherent surface structure. This redundancy not only increases memory and computation, but also means the representation neglects the underlying scene geometry, limiting robustness to novel viewpoints and edits. 

There is thus a strong incentive to develop representations that retain 3DGS’s photorealistic fidelity and speed while offering a simpler, more structured geometric forms.
Recent works have begun adding structural regularity to the Gaussian splatting paradigm. Scaffold-GS \cite{lu2024scaffold}
imposes a two-layer hierarchy of Gaussians anchored on a sparse grid, dynamically spawning primitives
only where needed to reduce overlap and redundancy. 
This approach yields a more compact model
and improves coverage of the scene while maintaining high rendering quality. 
On another front, methods like GSDF \cite{yu2024gsdf} and 3DGSR \cite{lyu20243dgsr}  integrate implicit surface representations into the 3DGS pipeline. 
By aligning Gaussians with a learned SDF field, these hybrid methods
enforce surface consistency and enable extraction of explicit meshes from the neural model. 
While effective in improving geometric fidelity, such approaches still retain a large number of Gaussian primitives and rely on complex networks or dual-branch optimizations for surface inference.
Moreover, simply constraining Gaussians to lie on a thin surface, essentially treating them as flat planes, has been shown to degrade rendering quality and produce incomplete surfaces. 
These limitations highlight the need for a fundamentally different strategy to achieve a \emph{simple}, \emph{structurally
regular}, and \emph{scalable} 3D representation.

In this paper, we introduce HaloGS~\footnote{The term “halo” serves as a metaphor for the volumetric ring of fine‑grained detail that gently envelops a compact, low‑poly geometry, as illustrated in Fig.~\ref{fig:pipeline}(a). Inspired by its name, `Ha' denotes our \emph{hybrid architecture}, while “lo” evokes \emph{loose coupling}, \emph{level‑of‑detail}, and \emph{low‑poly} geometry.}, a novel framework that decouples geometry and appearance using a dual representation. 
Instead of fusing shape and color into a single representation, \modelname models geometry with a set of learnable triangle primitives, while appearance is rendered via neural Gaussians that decoded from the triangles.  Benifiting from an adaptive training strategy, the two tasks receive mutual enhancement. Moreover, our custom CUDA-based triangle splatting kernel enables direct supervision of geometric primitives using predicted normals and depths from either diffusion models or appearance primitives, avoiding reliance on indirect supervision through rendering loss. Crucially, the triangle are selected to present geometry due to its more efficient and the ability to preserve shapeness and planeness.

Although HaloGS does not enforce 2D-manifold surfaces (i.e., via triangle soup), this loose connectivity already provides a natural basis for human‑made environments such as interiors and urban scenes. Rather than extracting dense meshes via Marching Cubes from implicit fields, we argue that simplicity and structural regularity are as important as raw accuracy for downstream tasks (e.g., collision detection, geometric editing, and real‑time streaming). To this end, we extract planar primitives at multiple levels of detail directly from our learnable triangles and stitch them into a compact mesh, preserving salient facets like walls, floors, and building facades while discarding redundant geometry, as shown in Fig.~\ref{fig:reaser} and~\ref{fig:comapct_mesh}. This compact mesh is more storage‑ and compute‑efficient than conventional dense reconstructions, yet remains highly expressive for both simple and complex scenes across indoor and outdoor scenarios.

%% file: sec/related_work.tex
\section{Related Works} 
Representing 3D scenes invariably demands a balance between compact, editable geometry and rich, view-dependent appearance. Below, we briefly review two lines of work that illustrate this trade-off: \vspace*{-2mm}
\paragraph{Neural Rendering \& Novel View Synthesis.}
Neural Radiance Fields (NeRF)~\cite{mildenhall2021nerf} brought photorealistic, view-dependent rendering by training MLPs to predict color and density along each ray. Despite their visual fidelity, NeRF incurs heavy per-ray sampling and embed geometry in an implicit manner, hindering scalability and direct editing. To accelerate inference, later efforts introduced explicit scene structures: point-based models~\cite{xu2022point}, sparse voxel or hash grids ~\cite{fridovich2022plenoxels, muller2022instant, xu2023grid}, and anisotropic 3D Gaussians~\cite{kerbl20233d}. Gaussian Splatting drives real-time rendering with quality on par with NeRF, and Scaffold-GS~\cite{lu2024scaffold} further boosts efficiency and detail. Urban-scale extensions~\cite{jiang2024horizon, liu2024citygaussian, ren2024octree, kerbl2024hierarchical} validate its scalability. 
However, these methods typically prioritize rendering capability at the expense of geometric rigidity, making it difficult to faithfully represent complex scenes with high‑frequency appearance variations without explicit geometry support. \vspace*{-2mm}

\paragraph{3D Reconstruction.} 
Reconstructing 3D geometry from multi-view images is a fundamental task in computer vision and computer graphics. Traditionally, images are first calibrated to generate dense point clouds, which are then optimized into an implicit field. Meshes are subsequently extracted from this field using the Marching Cubes \cite{lorensen1998marching} algorithm. Over the past decades, each stage of this pipeline has been extensively studied. In contrast, inspired by neural rendering techniques, recent works aim to bypass the intermediate point cloud and instead directly optimize an implicit field using rendering loss. For example, methods such as~\cite{wang2021neus, li2023neuralangelo, wang2023neus2, rosu2023permutosdf} integrate Signed Distance Functions (SDFs) into NeRF frameworks, while others~\cite{huang20242d, yu2024gaussian} optimize depth or opacity fields through Gaussian Splatting. Furthermore, two-branch frameworks~\cite{yu2024gsdf, lyu20243dgsr} have been proposed to jointly reconstruct differentiable geometry and appearance. However, all the aforementioned methods produce dense meshes that prioritize geometric accuracy but lack structural compactness. In contrast, compact meshes composed of large-area polygons are more controllable and easier to manipulate~\cite{chen2020bsp}. Similar to traditional dense reconstruction pipelines, compact mesh reconstruction from multi-view images also typically requires generating dense point clouds first, followed by shape detection methods~\cite{yu2022finding, li2019supervised, sharma2020parsenet} to approximate primitives such as planes, cylinders, spheres, and cones. These primitives are then converted into compact meshes~\cite{bauchet2020kinetic, Nan_2017_ICCV, sulzer2024concise}. Nevertheless, multi-view stereo (MVS) often produces noisy and incomplete point clouds, whereas our method directly reconstructs clean and complete triangle representations.

%% file: sec/method.tex
\begin{figure}[t]
  \centering
   \includegraphics[width=1\linewidth]
   {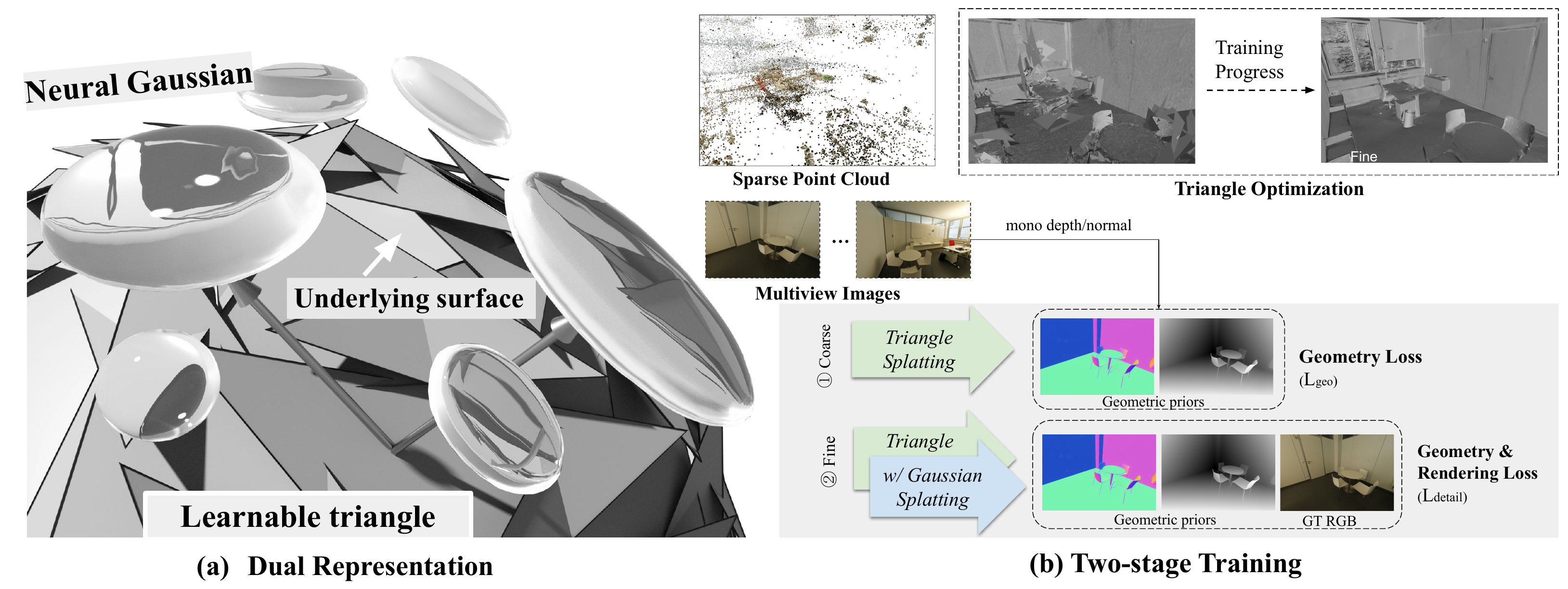}
   \vspace*{-4mm}
   \caption{\textbf{Overview of HaloGS.} Our proposed dual-representation is illustrated in (a), where learnable triangles explicitly fit the scene geometry, and neural Gaussians decoded from these triangles render the appearance. In (b), we depict our coarse-to-fine training strategy: during the coarse stage, monocular geometric priors supervise the positions and shapes of the triangles. Subsequently, in the fine stage, neural Gaussians decoded from these half-trained triangles are optimized using ground truth images. Concurrently, depth and normal maps rendered from the neural Gaussians provide additional refinement feedback to further enhance the triangle representation. }
   \vspace{-1.5em}
   \label{fig:pipeline}
\end{figure}

\section{Method}
\label{sec:method}
In this section, we first review the relevant preliminaries on representative Gaussian primitives that motivate our design. We then describe our dual representation, which combines triangle primitives with neural Gaussians, and continue with discussions of the technical details, as shown in Fig.~\ref{fig:pipeline}.
\subsection{Preliminaries}
\label{preliminaries}
3D Gaussian Splatting (3DGS)~\cite{kerbl20233d} represents a 3D scene with a set of anisotropic 3D Gaussian distributions, each of which is defined with its center $\bm{\mu}$ and covariance matrix $\bm{\Sigma}$ with
$G(\bm{x}) = e^{-\frac{1}{2} (\bm{x}-\bm{\mu})^T \bm{\Sigma}^{-1} (\bm{x}-\bm{\mu})}.$
In addition, each 3D Gaussian is also associated with a color $c$ and an opacity $\alpha$ for rendering. An efficient tile-based rasterizer is designed to render the 3D Gaussians by projecting them onto the image plane and blending their contributions.
Scaffold-GS~\cite{lu2024scaffold} later introduces a hierarchical architecture, where anchors are sparsely and uniformly distributed to preserve the geometric structure of the scene. Based on these anchors, anchor-wise features are decoded into the properties of neural Gaussians, which are then used for image rendering. Although Scaffold-GS achieves improved rendering quality and efficiency compared to 3DGS, the anchors themselves remain noisy and fail to represent meaningful geometry, as they are not directly optimized but are instead supervised through rendering loss. 
Based on Scaffold-GS, Octree-GS~\cite{ren2024octree} introduces a Level-of-Detail (LoD) based anchor structure, enabling real-time rendering of large scenes by selecting neural Gaussian during the rendering process. Additionally, the multi-scale arrangement of anchors provides higher fidelity for modeling scene details. During the training process, Octree-GS employs a progressive training strategy and improves upon the densification strategy of Scaffold-GS.

To improve geometric reconstruction, 2D Gaussian Splatting (2DGS)~\cite{huang20242d} squeeze 3D Gaussians to 2D Gaussians. In principle, "squeezing" Gaussians along one direction facilitates better alignment with potential surface geometry. Technically, each 2D Gaussian is defined on a local tangent plane in world space. The transformation matrix from a local coordinate to its
world-space counterpart is:
\begin{equation}
\mathbf{H} = \begin{bmatrix}\mathbf{s}_{u}\mathbf{t}_u&\mathbf{s}_{v}\mathbf{t}_v&0&\bm{\mu}\\0&0&0&1\end{bmatrix},
\label{H}
\end{equation}
where $\bm{\mu}$ denotes the center of a 2D Gaussian, while $(\mathbf{t}_u, \mathbf{t}_v)$ represent its two principal tangential directions, and $(\mathbf{s}_u, \mathbf{s}_v)$ are the corresponding scaling factors. 
In addition, intersection points between observing rays and 2D Gaussians are calculated following~\cite{sigg2006gpu}, before blending them for rendering.

\subsection{Representations}
\label{sec:representations}
HaloGS is a dual representation in which geometry and appearance are learned together in a loos-coupling manner.
As the name suggests, like a halo surrounding the center object.
We represent the scene surface as a collection of learned triangular primitives, where small neural Gaussians are attached to those triangles to encode view-dependent appearance. 
The triangle primitives provide an explicit, mesh‐like scaffold: they preserve sharp edges and coarse scene structure efficiently, which is difficult for purely volumetric schemes. 
In effect, the triangles capture low-frequency geometry, while the Gaussians parameterize high-frequency texture and lighting detail. This complementary design lets us exploit the best of both worlds: the triangle mesh ensures a compact, interpretable geometry (with far fewer parameters than a dense volumetric grid), while the Gaussian volumes provide photorealistic rendering, resolving the tension between qualities of geometric reconstruction and rendering that typically arises in a single-grain representation. 

\noindent \textbf{Learnable Triangle Primitives for Scene Geometry.}
We parameterize each triangle using its three movable vertices, $\{\mathbf{p}_0, \mathbf{p}_1, \mathbf{p}_2\}$. To rasterize the triangles, we follow the approach of 2DGS~\cite{huang20242d}, which defines a transformation matrix (Eq.~\ref{H}) that maps the coordinates of the intersection points from the local tangent space to the world space. Different from 2DGS, the transformation matrix $\mathbf{H}$ is now defined based on the triangle: 

\begin{equation}
\begin{aligned}
\left\{
\begin{array}{l}
\bm{\mu} = \dfrac{\mathbf{p}_0 + \mathbf{p}_1 + \mathbf{p}_2}{3}\\[8pt]
\mathbf{t}_u = \dfrac{\mathbf{p}_0 - \bm{\mu}}{\left\|\mathbf{p}_0 - \bm{\mu}\right\|_2}, \quad 
\mathbf{t}_v = \mathbf{n} \times \mathbf{t}_u \\[8pt]
\mathbf{s}_u = \left\|\mathbf{p}_0 - \bm{\mu}\right\|_2, \quad 
\mathbf{s}_v = \left| \mathbf{t}_v \cdot (\mathbf{p}_1 - \bm{\mu})\right|\\[8pt]
\end{array}
\right.
\quad \text{where} \quad
\mathbf{n} = \dfrac{(\mathbf{p}_1 - \mathbf{p}_0) \times (\mathbf{p}_2 - \mathbf{p}_0)}{\left\|(\mathbf{p}_1 - \mathbf{p}_0) \times (\mathbf{p}_2 - \mathbf{p}_0)\right\|_2}
\end{aligned},
\label{eq:local_triangle_frame}
\end{equation}
where the barycenter $\bm{\mu}$ of each triangle is set as the origin of the local coordinate frame, and the first vertex $\mathbf{p}_0$ always has a local coordinate $[1, 0]^T$ based on the above construction.

To enable triangle splatting, each triangle is also associated with a learnable opacity value $\alpha$. Additionally, following 3D Convex Splatting (3DCS)~\cite{held20243d}, we introduce two more learnable triangle-wise parameters, $\delta > 0$ and $\sigma > 0$, which are used to compute the contribution $w$ of each ray-triangle intersection point $\hat{\mathbf{x}}$ via an edge-preserving kernel:
\begin{equation}
w(\hat{\mathbf{x}})=\mathrm{Sigmoid}\left(-\sigma\log\left(\sum_{j=0}^2\exp\left(\delta~
\mathrm{dist}(\hat{\mathbf{x}}, \mathbf{e}_j)\right)\right)\right)\alpha, 
\label{eq:contribution}
\end{equation}
where $\mathrm{dist}(\hat{\mathbf{x}}, \mathbf{e}_j)$ is the Euclidean distance from $\mathbf{x}$ to triangle edge $\mathbf{e}_j$, and the intersection point $\hat{\mathbf{x}}$ between an observation ray of pixel $\mathbf{x}$ and a triangle is computed using the explicit ray-splat intersection algorithm~\cite{sigg2006gpu}. More details are provided in the supplementary.

Note that the triangles are optimized solely for geometric reconstruction. These triangle splats are then "rendered" into normal and depth maps to match the supervised targets through
\begin{equation}
\mathbf{N}(\mathbf{x}) = \sum_{i=1}^N \mathbf{n}_i w\left(\hat{\mathbf{x}}_i\right)  \prod_{j=1}^{i-1} \left(1 - w\left(\hat{\mathbf{x}}_j\right) \right), \quad
\mathbf{D}(\mathbf{x}) = \sum_{i=1}^N \mathbf{d}_i w\left(\hat{\mathbf{x}}_i\right)  \prod_{j=1}^{i-1} \left(1 - w\left(\hat{\mathbf{x}}_j\right) \right),
\label{eqn:NandD}
\end{equation}
where $\mathbf{d}_i$ denotes the distance from the $i$-th intersection point to the pixel in the world coordinate.
The $N$ ordered intersection points $\{\hat{\mathbf{x}}_i\}$ between the traingles and the pixel $\mathbf{x}$ are calculated using a CUDA-based rasterizer.

\noindent \textbf{Neural Gaussians for Scene Appearance.}
Inspired by Scaffold-GS \cite{lu2024scaffold}, we spawn neural Gaussians from the triangles for rendering. Specifically, each learnable triangle is further equipped with a local context feature $f_v\in\mathbb{R}^{32}$, a scaling factor $l_v\in\mathbb{R}^{3}$, and $k$ learnable offsets $O_v\in\mathbb{R}^{k \times 3}$, while the properties of neural Gaussians are decoded from them following~\cite{lu2024scaffold}. 
To further enhance rendering capability, we embed these Gaussians in an octree structure following~\cite{ren2024octree}, as elaborate in Sec.~\ref{sec:octreegs}.

\subsection{Training and Losses}
\label{geo_training}
To stabilize the training and also improve the robustness for complex real‑world scenes, we adopt a two-stage, coarse‑to‑fine strategy (see Fig.~\ref{fig:pipeline}).
Technically, we begin with a sparse point cloud, typically obtained with COLMAP~\cite{schoenberger2016sfm} following common practices, and then initialize each triangle by placing three vertices $(\mathbf{p}_0, \mathbf{p}_1, \mathbf{p}_2)$ randomly on a small circle centered at each SfM point.

\noindent \textbf{Coarse-to-Fine Strategy.}
The coarse-stage training guides all initialized triangles towards reconstructing accurate geometry by leveraging two monocular priors: a depth map \(\mathbf{D}_{\mathrm{ref}}\) predicted by Depth‑Anything‑V2~\cite{yang2024depth} and a normal map \(\mathbf{N}_{\mathrm{ref}}\) from StableNormal~\cite{ye2024stablenormal}.
These priors supervise each triangle's position through a geometric loss that penalizes deviations from the predicted depth and normals based on Eq.~\ref{eqn:NandD}
\begin{equation}
\mathcal{L}_{\text{geo}}(\mathbf{D}, \mathbf{N}; \mathbf{D}_{\text{ref}}, \mathbf{N}_{\text{ref}})= \lambda_\text{d} \|\mathbf{D}-\mathbf{D}_\text{ref}\|_1 + \|\mathbf{N}-\mathbf{N}_\text{ref}\|_1 + \|1 - \mathbf{N}^{\top}\mathbf{N}_\text{ref}\|_1,
\end{equation}
where $\lambda_\text{d}$ is a user-prescribed weight. To ensure metric consistency, the monocular depth predictions are calibrated using RANSAC with sparse SfM points, following the method in~\cite{kerbl2024hierarchical}.

The fine-stage training continues to refine geometric details using neural Gaussians, which both enhances accuracy through rendering‑based supervision and mitigates compounded errors in the monocular geometry priors.
Following the design in~\cite{lu2024scaffold}, we decode neural Gaussians from each learnable triangle and supervise them using the ground‑truth images. The optimized Gaussians render a finer-detailed depth map \(\mathbf{D}_{\text{gs}}\) and normal map \(\mathbf{N}_{\text{gs}}\), which then supervise the triangle primitives through the loss:
\begin{equation}
\scalebox{0.93}{$
\begin{aligned}
\mathcal{L}_{\mathrm{detail}}
&=
\mathcal{L}_{\text{geo}}(\mathbf{W}\!\odot \!\mathbf{D}, \mathbf{W}\!\odot\!\mathbf{N}; \mathbf{W}\!\odot\!\mathbf{D}_{\text{gs}}, \mathbf{W}\!\odot \!\mathbf{D}_{\text{gs}}) + \mathcal{L}_{\mathrm{geo}}\bigl(
   \mathbb{W}\!\odot \!\mathbf{D},\,
    \mathbb{W}\!\odot\!\mathbf{N}
   \,;\,
   \mathbb{W}\!\odot\!\mathbf{D}_{\mathrm{ref}},\,
   \mathbb{W}\!\odot\! \mathbf{N}_{\mathrm{ref}}
\bigr) \\ 
&+ \lambda_{\mathrm{rgb}}\Bigl(
   (1-\lambda_{\mathrm{c}})  \|\mathbf{C}_{\mathrm{gt}}-\mathbf{C}_{\mathrm{gs}}\|_{1}
   + \lambda_{\mathrm{c}} \text{SSIM}\bigl(\mathbf{C}_{\mathrm{gt}},\mathbf{C}_{\mathrm{gs}}\bigr)
   + \lambda_{\mathrm{s}} \mathcal{L}_{scaling}\Bigr).
\end{aligned}$}
\end{equation}
Here, we introduce a weight map \(\mathbf{W}\) (and its complement $\mathbb{W} \!=\! \mathbf{1}\!-\!\mathbf{W}$) obtained via a wavelet transform on the RGB images, to enhance high‑frequency regions (i.e., its value approaches 1 on edges and in texture areas, and 0 elsewhere). $\mathcal{L}_{scaling}$ is a volume regularization from Scaffold-GS \cite{lu2024scaffold}. The weights $\lambda_{\mathrm{c}}$, $\lambda_{\mathrm{s}}$, $\lambda_{rgb}$ and $\lambda_d$ are set to $0.2$, $0.01$, $10$ and $10$ in our experiments. 

\noindent \textbf{Adaptive Density Control.} 
To obtain complete, clean reconstructions even from low‑quality initializations, we introduce an adaptive density control algorithm comprising splitting and pruning operations.
First, we detect edges requiring refinement by computing, for each edge \(e=(\mathbf{p}_0,\mathbf{p}_1)\), the directional‑derivative difference
\begin{equation}
\nabla_{e} \;=\; \bigl(\nabla f(\mathbf{p}_1) - \nabla f(\mathbf{p}_0)\bigr)\cdot\frac{(\mathbf{p}_1 - \mathbf{p}_0)}{\|\mathbf{p}_1 - \mathbf{p}_0\|_2},
\end{equation}
where \(\nabla f(\mathbf{p})\) is the spatial gradient of the implicit field represented by 3DGS at point \(\mathbf{p}\). If \(\nabla_{e}\) exceeds a user‑defined threshold (set to \(1\times10^{-5}\) in our experiments), we split the longest such edge by inserting a new vertex at its midpoint, bisecting the triangle.
Next, we prune triangles that contribute little to the final model. We measure each triangle’s contribution by its visibility across views and its opacity; those with low observation or negligible opacity are removed.
To enforce a clear binary opacity so that triangles become either fully opaque or fully transparent, we add an entropy loss on each triangle’s opacity \(\alpha\), following~\cite{guedon2024sugar}. 
More details are provided in the supplementary.

\subsection{Level-of-Detail Planar Abstraction}
\label{sec:octreegs}
\noindent \textbf{Enhanced Neural Rendering.}
To further leverage the explicit nature of learnable triangles and enable more structured scene management, 
we construct a multi-scale Level-of-Detail (LoD) planar representation. We first approximate planar regions at successive detail levels by clustering the optimized triangles using GoCoPP~\cite{yu2022finding}. Each set of planar primitives then seeds an Octree‑GS~\cite{ren2024octree} hierarchy, enabling LoD-aware rendering that delivers enhanced rendering fidelity.  

\noindent \textbf{Enhanced Geometric Reconstruction.}
\label{compactmesh}
The extracted planar primitives serve not only as initialization for Octree‑GS but also as building blocks for a compact mesh. We stitch these planes into a coherent surface using kinetic shape reconstruction~\cite{bauchet2020kinetic} or concise plane arrangements\cite{sulzer2024concise}. 
Both approaches require an oriented point cloud: we therefore sample points uniformly from the optimized triangles and assign each point the triangle’s normal. 
Implementation details are provided in the supplementary.  

%% file: sec/experiments.tex
\section{Experiments}
\label{sec:exp}

\begin{figure*}[t!]
\centering
\includegraphics[width=\linewidth]{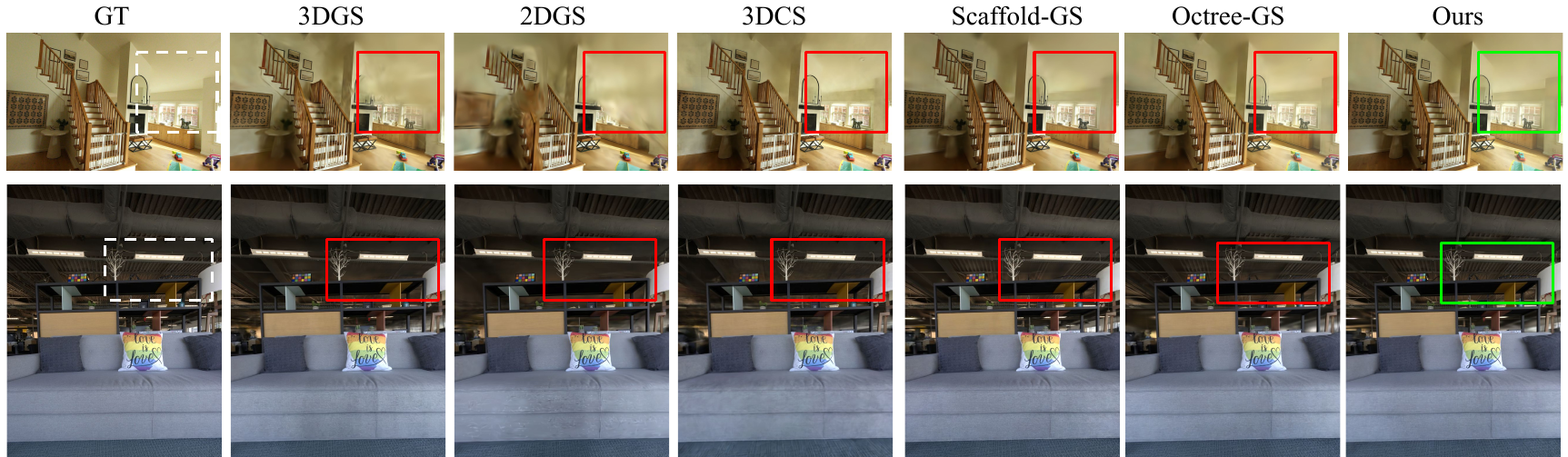}
\caption{
We evaluate our method against state-of-the-art approaches~\cite{kerbl20233d,huang20242d,held20243d,lu2024scaffold, ren2024octree} on challenging Zip-NeRF and VR-NeRF scenes that span both expansive layouts and intricate details. Colored patches draw attention to areas where our approach excels, faithfully reconstructing fine structures and complex planar surfaces, such as wall-mounted mirrors and intricate ceiling ornaments, which existing baselines struggle to capture.}
\label{fig:Rendering_comparison}
\centering
\vspace{-1em}
\end{figure*}

\begin{figure}[t]
  \centering
  \vspace{-1em}
\includegraphics[width=1\linewidth]{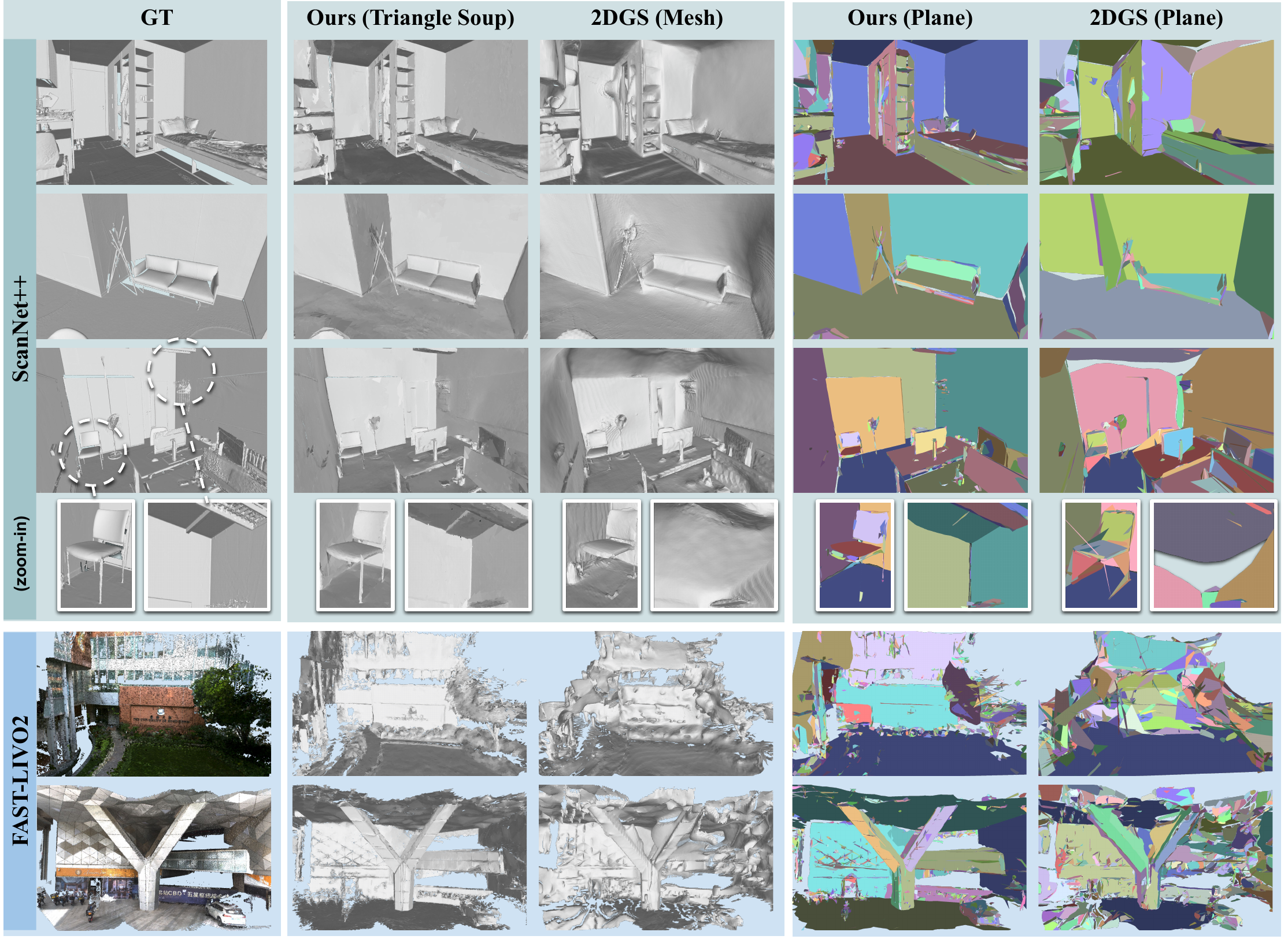}
   \caption{\textbf{Geometric reconstruction comparison.} 
   We visualize the learned geometric representations our Triangle Soup, 2DGS meshes, our extracted planes, and 2DGS planes on two representative datasets. 
   The top four rows show ScanNet++ indoor scenes with available ground‑truth meshes; insets highlight fine structural details. The bottom two rows present results on FAST‑LIVO2 outdoor captures with ground‑truth point clouds. 
   Our Triangle Soup faithfully preserves geometry fidelity, capturing sharp edges and fine details.
   }
   
   \label{fig:geo_compare} 
\end{figure}

\begin{figure}[t]
  \centering
\includegraphics[width=1\linewidth]{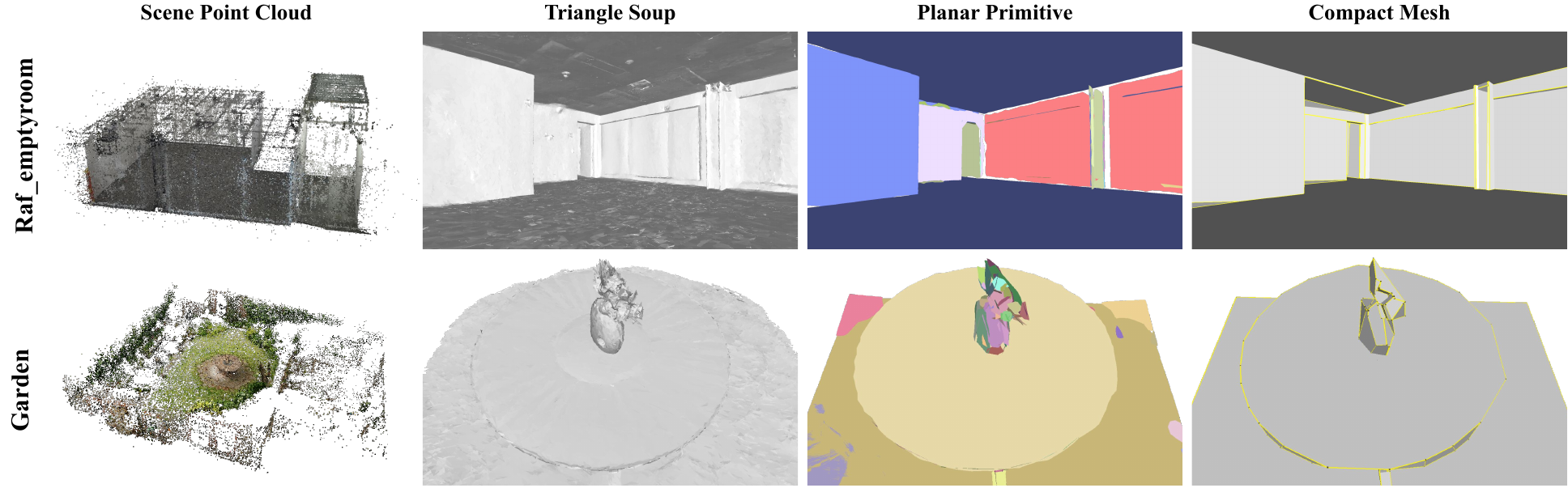}
   \caption{
   \textbf{Visualization of the compact mesh.} We present two representative scenes: Raf\_emptyroom (from VR‑NeRF) and Garden (from MipNeRF‑360) in both indoor and outdoor settings.
   }
   \label{fig:comapct_mesh}
   \vspace{-1.em}
\end{figure}

\subsection{Experimental Setup}

\noindent \textbf{Datasets.} 
We evaluate HaloGS on 32 real‑world scenes drawn from diverse benchmarks: two from DeepBlending~\cite{hedman2018deep}, ten from ScanNet++~\cite{yeshwanth2023ScanNet++}, eleven from VR‑NeRF~\cite{VRNeRF}, four from Zip‑NeRF~\cite{barron2023zipnerf}, one from Matrix‑City~\cite{li2023matrixcity}, and four from FAST‑LIVO2~\cite{zheng2024fast}, spanning a broad spectrum of indoor and outdoor environments.

\noindent\textbf{Implementation Details.}  
\label{exp:imple}
Our pipeline consists of a coarse stage (10k iterations) followed by a fine stage (10k iterations). During training, we extract LoD planar shapes at 10k, 15k, and 20k iterations, which are then used to initialize Octree-GS. The Octree-GS model undergoes 30k iterations for training. All experiments are conducted on a single NVIDIA A800 GPU with 80G memory. More implementation details are provided in the supplementary material.    

\textbf{Baselines.}
We compare our method against state-of-the-art approaches in both novel view synthesis and geometry reconstruction. For rendering comparisons, we consider 3DGS~\cite{kerbl20233d}, 2DGS~\cite{huang20242d}, 3DCS~\cite{held20243d}, Scaffold-GS~\cite{lu2024scaffold}, and Octree-GS~\cite{ren2024octree}. All methods are trained for 40k iterations to ensure fair comparison under consistent settings. For geometry evaluation, we primarily compare our method with MVS~\cite{schoenberger2016mvs}, 2DGS~\cite{huang20242d}, and GFSGS~\cite{jiang2024geometry}. Unless otherwise stated, all evaluations are conducted using 1/8 of the images as the test set and the remaining 7/8 for training across all datasets.

\textbf{Quantitative Metrics.}
We evaluate both rendering fidelity and geometric compactness here. For rendering, we report the widely-used PSNR, SSIM~\cite{wang2004image} and LPIPS~\cite{zhang2018unreasonable}. To assess geometry, we follow the protocol of ~\cite{yu2022finding}: we extract planar primitives and measure their fidelity via the Chamfer Distance to the ground-truth mesh, and their simplicity by the total number of extracted planes. Below, we highlight the \textbf{best} and \underline{second-best} results in the reported tables.

\subsection{Results Analysis}

\noindent \textbf{Rendering Comparisons.} 
Our method retained high rendering quality compared to state-of-the-art 3DGS-based methods across various indoor and outdoor datasets, as shown in Table \ref{tab:render_1} and \ref{tab:render_2}. 
Our approach demonstrated exceptional performance in achieving superior rendering quality in texture-less and low-light regions, as illustrated in Fig. \ref{fig:Rendering_comparison}. Specifically, in these areas, the sparse and inaccurate SfM points often lead to suboptimal outcomes with baseline methods. Nonetheless, our method effectively tackled this challenge by leveraging its precise and consistent geometric model.

\noindent \textbf{Geometry Comparisons.}
To evaluate the reconstructed geometry, we conduct comprehensive evaluations on two complementary benchmarks: ScanNet++ \cite{yeshwanth2023ScanNet++} for indoor scene reconstruction and FAST-LIVO2 \cite{zheng2024fast} for outdoor environments. As the baselines typically output point clouds or dense meshes, we additionally extract LoD planar shapes from their results using the same extraction strategy and parameters as applied to ours for fair comparison. As quantified in Table~\ref{tab:geo_compare}, our method achieves better geometric accuracy with lower Chamfer distance while maintaining superior structural compactness.
Our triangle soup representation preserves planar regularity and sharp geometric features as illustrated in Fig.~\ref{fig:geo_compare}. 
With our adaptive triangle splitting strategy, the representation  
achieves accuracy on flat surfaces while maintaining smooth continuity across curved regions.
This geometrically faithful foundation supports a compact planar abstraction that aligns with ground truth. 
In contrast, 2DGS \cite{huang20242d} suffers from surface over-smoothing and lack of details due to its depth-fusion mesh extraction, exacerbated by coupled geometry-appearance optimization. This entanglement causes models overfitting to distort geometric estimation, resulting in misaligned planar boundaries and improper surface connections in occluded regions.

\noindent \textbf{Application to Compact Mesh Reconstruction.}
Starting from the optimized triangles, we reconstruct compact mesh following the description in Sec.~\ref{compactmesh}. Fig.~\ref{fig:comapct_mesh} and Fig.~\ref{fig:reaser} visualize the resulting mesh alongside the triangle soup and planar primitives, highlighting how our compact meshes faithfully reproduce geometry.

\begin{table}[t!]
\centering
\renewcommand{\arraystretch}{1.15}
\setlength{\tabcolsep}{1pt}
   \caption{\textbf{Rendering comparisons against baselines} over three indoor datasets~\cite{hedman2018deep, yeshwanth2023ScanNet++, VRNeRF}. Our method is compared with 3DGS~\cite{kerbl20233d}, 2DGS~\cite{huang20242d}, 3DCS~\cite{held20243d}, Scaffold-GS~\cite{lu2024scaffold}, and Octree-GS~\cite{ren2024octree}. Our method consistently outperforms baselines in rendering quality. We report visual quality metrics, per-frame averages for Gaussians rendered \textit{\#Render} and average storage size \textit{\#Mem}.
   }
\label{tab:render_1}
\resizebox{1\linewidth}{!}{
\begin{tabular}{c|cccc|cccc|cccc}
\toprule
Dataset & \multicolumn{4}{c|}{Deep Blending} & \multicolumn{4}{c|}{ScanNet++} & \multicolumn{4}{c}{VR-NeRF} \\
\begin{tabular}{c|c} Method & Metrics \end{tabular}  & PSNR\(\uparrow\) & SSIM\(\uparrow\) & LPIPS\(\downarrow\) & \#Render/\#Mem & PSNR\(\uparrow\) & SSIM\(\uparrow\) & LPIPS\(\downarrow\) & \#Render/\#Mem & PSNR\(\uparrow\) & SSIM\(\uparrow\) & LPIPS\(\downarrow\) & \#Render/\#Mem \\
\midrule
			
3DGS & 29.46 & 0.903 & 0.242 & 398K/ 705.6M & 30.43 & \underline{0.930} & 0.157 & 327K/ 229.1M & 30.20 & \underline{0.918} & \underline{0.189} & 457K/ 547.8M \\
           
2DGS & 29.32 & 0.899 & 0.257 & 196K/ 335.3M & 30.06 & 0.929 & 0.168 & \underline{170K}/ 118.8M & 29.31 & 0.899 & 0.246 & \textbf{144K}/ \textbf{165.6M} \\
         
3DCS & 29.63 & 0.901 & \underline{0.236} & \underline{150K}/ 287.1M & \textbf{30.59} & 0.929 & 0.157 & \textbf{105K}/ \textbf{80.0M} & \underline{30.64} & 0.908 & 0.211 & \underline{264K}/ 333.8M  \\

Scaffold-GS & \underline{30.24} & \underline{0.909} & 0.239 & 207K/ \underline{125.5M} & 28.6 & 0.923 & 0.159 & 195K/ \underline{87.6M} & 30.37 & \textbf{0.925} & \textbf{0.161} & 850K/ 498.7M \\

Octree-GS & 30.18 & \underline{0.909} & 0.243 & \textbf{102K}/ \textbf{118.6M} & 30.27 & \underline{0.930} & \underline{0.154} & 268K/ 170.6M & 30.12 & 0.912 & 0.200 & 269K/ 387.3M \\

\hline

Ours & \textbf{30.31} & \textbf{0.911} & \textbf{0.226} & 193K/ 129.8M & \underline{30.51} & \textbf{0.931} & \textbf{0.149} & 336K/ 102.0M & \textbf{30.84} & 0.916 & 0.201 & 428K/ \underline{198.2M} \\
\bottomrule
\end{tabular}}
\end{table}

\begin{table}[t!]
\vspace{-1.5em}
\centering
\renewcommand{\arraystretch}{1.15}
\setlength{\tabcolsep}{1pt}
   \caption{\textbf{Rendering comparisons against baselines} over an indoor-outdoor dataset~\cite{barron2023zipnerf} and two outdoor datasets~\cite{li2023matrixcity,zheng2024fast}. Our method outperforms baselines in rendering quality. We report visual quality metrics, per-frame averages for Gaussians rendered \textit{\#Render} and average storage size \textit{\#Mem}.
   }
\label{tab:render_2}
\resizebox{1\linewidth}{!}{
\begin{tabular}{c|cccc|cccc|cccc}
\toprule
Dataset & \multicolumn{4}{c|}{Zip-NeRF} & \multicolumn{4}{c|}{Matrix-City} & \multicolumn{4}{c}{FAST-LIVO2} \\
\begin{tabular}{c|c} Method & Metrics \end{tabular}  & PSNR\(\uparrow\) & SSIM\(\uparrow\) & LPIPS\(\downarrow\) & \#Render/\#Mem & PSNR\(\uparrow\) & SSIM\(\uparrow\) & LPIPS\(\downarrow\) & \#Render/\#Mem & PSNR\(\uparrow\) & SSIM\(\uparrow\) & LPIPS\(\downarrow\) & \#Render/\#Mem \\
\midrule

3DGS & 25.39 & 0.811 & 0.360 & 243K/ 229.5M & 26.79 & 0.821 & 0.241 & 799K/ 3419.4M & \textbf{29.96} & 0.877 & 0.156 & 520K/ 760.2M \\

2DGS & 23.91 & 0.775 & 0.411 & \textbf{112K}/ \textbf{102.5M} & 25.76 & 0.796 & 0.308 & 327K/ 1071.0M & 28.38 & 0.852 & 0.201 & 448K/ 661.0M \\

3DCS & \underline{25.71} & \underline{0.817} & \textbf{0.334} & 378K/ 392.2M & 27.70 & 0.834 & 0.238 & \textbf{187K}/ 689.6M & 29.85 & 0.875 & 0.147 & \textbf{217K}/ 226.2M \\

Scaffold-GS & 24.13 & 0.806 & 0.354 & 317K/ \underline{129.5M} & 27.45 & \textbf{0.864} & 0.212 & 317K/ \textbf{317.8M} & 27.15 & 0.827 & 0.167 & \underline{400K}/ 225.2M \\

Octree-GS & 25.31 & 0.813 & 0.357 & \underline{201K}/ 187.8M & \underline{28.85} & 0.859 & \underline{0.202} & \underline{252K}/ \underline{391.5M} & \underline{29.92} & \underline{0.882} & \underline{0.139} & 457K/ \textbf{174.9M} \\

\hline

Ours & \textbf{25.73} & \textbf{0.818} & \underline{0.337} & 413K/ 213.3M & \textbf{28.86} & \underline{0.862} & \textbf{0.192} & 405K/ 474.5M & 29.81 & \textbf{0.883} & \textbf{0.137} & 497K/ \underline{202.1M} \\
\bottomrule
\end{tabular}}
\end{table}

\begin{table}[t!]
\centering
\caption{\textbf{Geometry comparisons.} We compare our method against other baselines in the planar shape extraction setting. Specifically, the point clouds or dense meshes generated by the baseline methods are used as input to the planar extraction method~\cite{yu2022finding}. We evaluate the fidelity score by computing the Chamfer distance between the extracted planar shapes and the ground truth geometry, and assess the simplicity score by counting the number of planar primitives.
}
\label{tab:geo_compare}
\resizebox{0.8\linewidth}{!}{
\begin{tabular}{c|cc|cc}
\toprule
\multicolumn{1}{c|}{Dataset} & \multicolumn{2}{c|}{ScanNet++} & \multicolumn{2}{c}{FAST-LIVO2} \\
\multicolumn{1}{c|}{Method \& Metrics} & Ch\_L2(cm) \(\downarrow\) & \# Planar Primitives \(\downarrow\) & Ch\_L2(cm) \(\downarrow\) & \# Planar Primitives \(\downarrow\) \\
\midrule
\multicolumn{1}{c|}{MVS} & 17.00 & 9658.70 & 31.68 & 5184.50 \\
\multicolumn{1}{c|}{2DGS} & 11.25 & 3967.60 & 41.67 & \textbf{2678.50} \\
\multicolumn{1}{c|}{GFSGS} & 17.29 & 10649.90 & 26.37 & 3438.00\\
\multicolumn{1}{c|}{Ours} & \textbf{7.84} & \textbf{1966.10} & \textbf{26.06} & 3637.25 \\
\bottomrule
\end{tabular}}
\end{table}

\begin{table}[t!]
\centering
\caption{\textbf{Quantitative Results on Ablation Studies.} }
\label{tab:ablation}
\resizebox{\linewidth}{!}{
\begin{tabular}{l|cccc|cccc}
\toprule
Scene & \multicolumn{4}{c|}{Playroom} & \multicolumn{4}{c}{Dr Johnson} \\
\begin{tabular}{c|c} Method & Metrics \end{tabular}  & PSNR \(\uparrow\) & SSIM \(\uparrow\) & LPIPS \(\downarrow\) & \# Triangles \(\downarrow\) & PSNR \(\uparrow\) & SSIM \(\uparrow\) & LPIPS \(\downarrow\)  & \# Triangles \(\downarrow\) \\
\midrule
\text{HaloGS (Full)} & \textbf{30.86} & \textbf{0.913} & 0.229 & 166K & \textbf{29.75} & \textbf{0.909} & 0.223 & 153K \\ \hline
w/o enhancing geomtirc detail using Gaussians & 30.57 & 0.910 & 0.234& 150K & 29.69 & 0.907 &  0.225 & 146K \\
w/o triangle pruning  &30.62 & 0.910 & \textbf{0.205}& 594K & 29.67&  0.907 & \textbf{0.215} & 708K \\ 
w/o adaptive density control of triangles & 30.65 & 0.911 & 0.230 & \textbf{36K} & 29.66 & 0.908 & 0.234 & \textbf{80K} \\
\bottomrule
\end{tabular}}
\end{table}

\subsection{Ablation Studies}

\begin{figure}[t]
  \centering
   \includegraphics[width=1\linewidth]{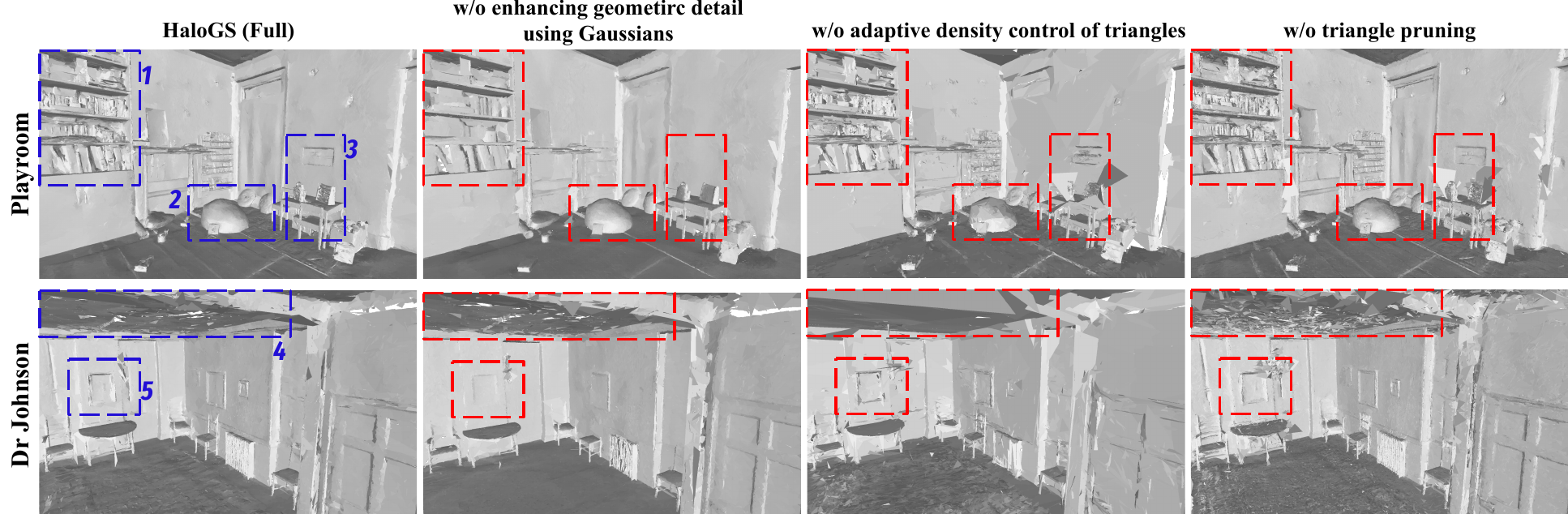}
   \caption{\textbf{Ablation results.}  Visualizations of triangle soup on DB~\cite{hedman2018deep} ablation studies. Numbered insets highlight where each omission degrades reconstruction quality.
   }
   \vspace*{-4mm}
   \label{fig:ablation}
   
\end{figure}

In this section, we examine the effectiveness of each individual module. Quantitative metrics and qualitative visualizations are provided in Table \ref{tab:ablation} and Fig. \ref{fig:ablation}.

\noindent \textbf{Geometric Detail Enhancement Using Neural Gaussians.} To evaluate the benefits of the neural Gaussians to the learnable triangles in the coarse-to-fine geometric optimization (Sec. \ref{geo_training}), we conduct an ablation study by disabling neural Gaussian-based refinement. The degraded configuration exhibits significant structural detail loss, particularly in bookshelves, tabletops, and wall-mounted areas (Fig. \ref{fig:ablation}, Column 2, Patches 1/3/5). These results demonstrate that augmenting geometry learning with rendering results leveraging neural Gaussians is essential for recovering fine detail.

\noindent \textbf{Adaptive Density Control of Triangles.} 
To evaluate the adaptive density control of triangles, we disable it and trained the model solely from the SfM-initialized point cloud. While the optimized triangles coarsely capture the overall scene structure, they exhibit significant noise and lack fine detail, as illustrated in Fig.~\ref{fig:ablation}, Column 3. This degradation is primarily due to the poor initialization inherent to SfM points.

\noindent \textbf{Triangle Pruning.} 
We further disabled our triangle pruning mechanism, which eliminates triangles based on view-space contribution and opacity. Without this regulation, three critical artifacts emerge: (1) significant redundancy in primitive accumulation (+311\% triangle count, see Table \ref{tab:ablation}), (2) unconstrained, oversized triangles, and (3) geometric outliers in occluded regions (Fig. \ref{fig:ablation}, Column 2, Patches 3 and 4). These experimental findings confirm that our pruning mechanism effectively ensures geometric accuracy while maintaining a compact model representation.

%% file: sec/conclusion.tex
\section{Conclusion}
\label{sec:conclusion}
HaloGS introduces a novel hybrid representation that loosely couples compact low‑poly geometry with Gaussian splats, uniting structural compactness and photorealistic appearance. Through the planar‑primitive fitting and mesh extraction pipeline, HaloGS maximally compresses primitive counts while preserving visual fidelity and enabling efficient rendering. Moreover, each intermediate output, such as segmented planar regions, can be repurposed for downstream tasks like reflection modeling, collision detection, and semantic reasoning. While HaloGS excels on large planar surfaces common in indoor and urban scenes, it can be challenged by highly curved or intricate geometries and exhibits sensitivity to the chosen coarse‑to‑fine training schedule (see Supplementary for a detailed analysis). We envision extending this work with higher‑order primitives and hierarchical fitting to better capture non‑planar structures. By bridging classical geometry representations and advanced immersive rendering, HaloGS paves the way for a wide range of applications, including AR/VR content generation, real‑time robotics mapping, and embodied AI, where efficiency, compactness, and visual quality are paramount.

%% file: sec/supplementary.tex
\newpage
\section{Supplementary Material}
The following sections are organized as:
\begin{itemize}
    \item \textbf{Sec.~\ref{sec:impl_details}} details the implementation of HaloGS, including the differentiable triangle rasterizer, adaptive density control strategy, normal orientation technique, and some configurations.
    \item \textbf{Sec.~\ref{sec:moreexp}} presents additional experimental results along with per-scene metric values.
    \item \textbf{Sec.~\ref{sec:limitation}} discusses the limitations of our approach and outlines directions for future works.
\end{itemize}

\subsection{Implementation Details}
\label{sec:impl_details}
 
\subsubsection{Differentiable Triangle Rasterizer} 
As discussed in Sec.~\ref{sec:representations}, we implement an efficient differentiable triangle rasterizer that enables direct supervision of triangles using predicted normals and depths derived from either diffusion models or appearance primitives. In this section, we provide further details on the forward rendering and backpropagation processes.
\begin{figure}[!ht]
  \centering
\includegraphics[width=1\linewidth]{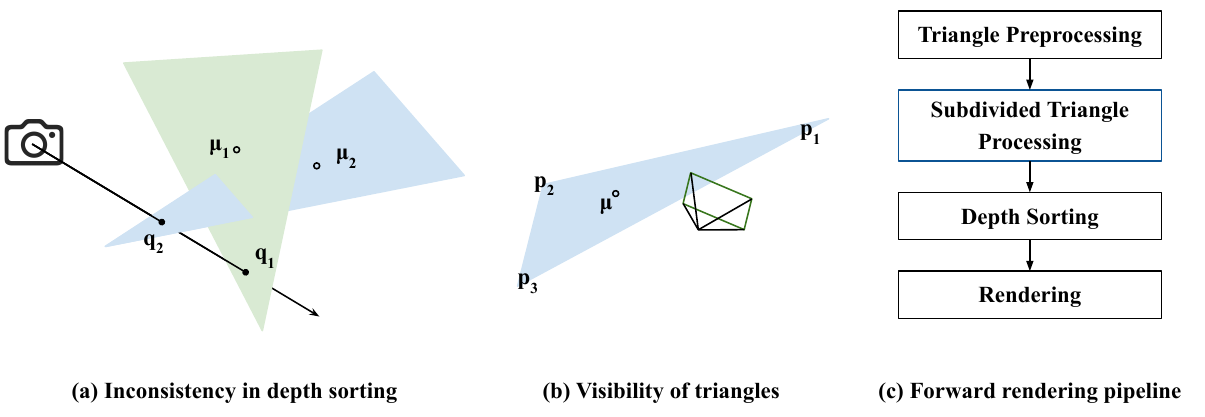 }
   \vspace*{-6mm}
   \caption{
   \textbf{ Two issues in forward rendering and our rendering pipeline.}
   }
   \label{fig:forward_rendering_pipeline}
\end{figure}

\noindent \textbf{Forward Rendering Process.} To achieve unbiased depth map rendering, we adopt the explicit ray-splat intersection algorithm~\cite{sigg2006gpu}, similar to 2DGS~\cite{huang20242d}, which determines the corresponding planar coordinate of each screen pixel and computes the rendering contribution on the local 2D tangent plane. This contrasts with 3DCS~\cite{held20243d}, where the contribution is calculated directly on the image plane. By employing triangle primitives and introducing an edge-preserving contribution function (as defined in Eq.~3 of the main text), two potential issues may arise:

\textit{Issue 1.} Before computing ray-triangle intersections, visible triangles are typically sorted for each camera based on their barycenters. However, for large triangles, a significant discrepancy may exist between the barycenter and the actual ray-triangle intersections, leading to depth sorting inaccuracies, as illustrated in Fig.~\ref{fig:forward_rendering_pipeline}(a). 

\textit{Issue 2.} Relying solely on a triangle’s barycenter to determine visibility can be insufficient. As shown in Fig.~\ref{fig:forward_rendering_pipeline}(b), vertex $\mathbf{p}_{1}$ is visible, while vertices $\mathbf{p}_{2}$, $\mathbf{p}_{3}$, and the barycenter $\bm{\mu}$ are occluded. 

We implement triangle subdivision to address \textit{Issue 1} and establish a new criterion for visibility determination to resolve \textit{Issue 2}. The forward rendering pipeline is illustrated in Fig.~\ref{fig:forward_rendering_pipeline}(c).

\textbf{1) Triangle Preprocessing.} We implement a custom CUDA kernel to process triangle soup. A triangle is considered visible if at least one of its three vertices is visible. For each visible triangle, we define a local coordinate frame (Eq.~2 in the main text) and compute the transformation matrix $\mathbf{H}$ (Eq.~1 in the main text). To address \textit{Issue 1}, we introduce triangle subdivision to reduce the discrepancy between ray-triangle intersection points and the triangle barycenters, thereby resolving inconsistencies in depth sorting. Subdivision is applied recursively along the edges of visible triangles until all edges of each subdivision triangle are shorter than a predefined threshold $\delta$. Each subdivision triangle inherits the ID of its original parent triangle. 

\textbf{2) Subdivided Triangle Processing.} We further process the subdivided triangles by first validating their visibility: a subdivided triangle is marked as visible if at least one of its three vertices is visible. For each visible subdivided triangle, we project its vertices onto the image plane to compute the number of tiles it overlaps. We also record the view-space depth of each subdivided triangle, computed using its barycenter, for the subsequent depth sorting. 

\textbf{3) Depth Sorting.} Following the approach of 2DGS~\cite{huang20242d}, we assign each subdivided triangle a sorting key that encodes its view-space depth and the ID of the tile it overlaps. We then perform an efficient GPU-based Radix sort to order the subdivided triangles based on these keys. 

\textbf{4) Rendering.} During rendering, while subdivided triangles are used for depth sorting, intersection calculations are performed on the original triangles. Following the 2DGS~\cite{huang20242d} approach, we define the camera ray corresponding to a pixel $\mathbf{x} = (x, y)^T$ as the intersection of two 4D homogeneous planes, $\mathbf{h}_x = (-1, 0, 0, x)^T$ and $\mathbf{h}_y = (0, -1, 0, y)^T$. These planes are then transformed into the local triangle coordinate using the transformation matrix $\mathbf{H}$: 

\begin{equation}
    \mathbf{h}_u=(\mathbf{W}\mathbf{H})^T\mathbf{h}_x, \quad\mathbf{h}_v=(\mathbf{W}\mathbf{H})^T\mathbf{h}_y,
\end{equation}
where $\mathbf{W}$ denotes the transformation matrix from world space to screen space. The computation of the triangle-ray intersection $\hat{\mathbf{x}}$ in the local triangle coordinate is given by: 
\begin{equation}
\hat{\mathbf{x}}=\left(\frac{\mathbf{h}_u^2\mathbf{h}_v^4-\mathbf{h}_u^4\mathbf{h}_v^2}{\mathbf{h}_u^1\mathbf{h}_v^2-\mathbf{h}_u^2\mathbf{h}_v^1}, \frac{\mathbf{h}_u^4\mathbf{h}_v^1-\mathbf{h}_u^1\mathbf{h}_v^4}{\mathbf{h}_u^1\mathbf{h}_v^2-\mathbf{h}_u^2\mathbf{h}_v^1}\right),
\end{equation}
where $\mathbf{h}_u^i$ and $\mathbf{h}_v^i$ denote the $i$-th value of the 4D homogeneous plane parameters. Subsequently, the rendering contribution is computed using Eq. 3 in the main text. By following Eq. 4 in the main text, we can render depth and normal images. 

\begin{figure}[!ht]
  \centering
\includegraphics[width=1\linewidth]{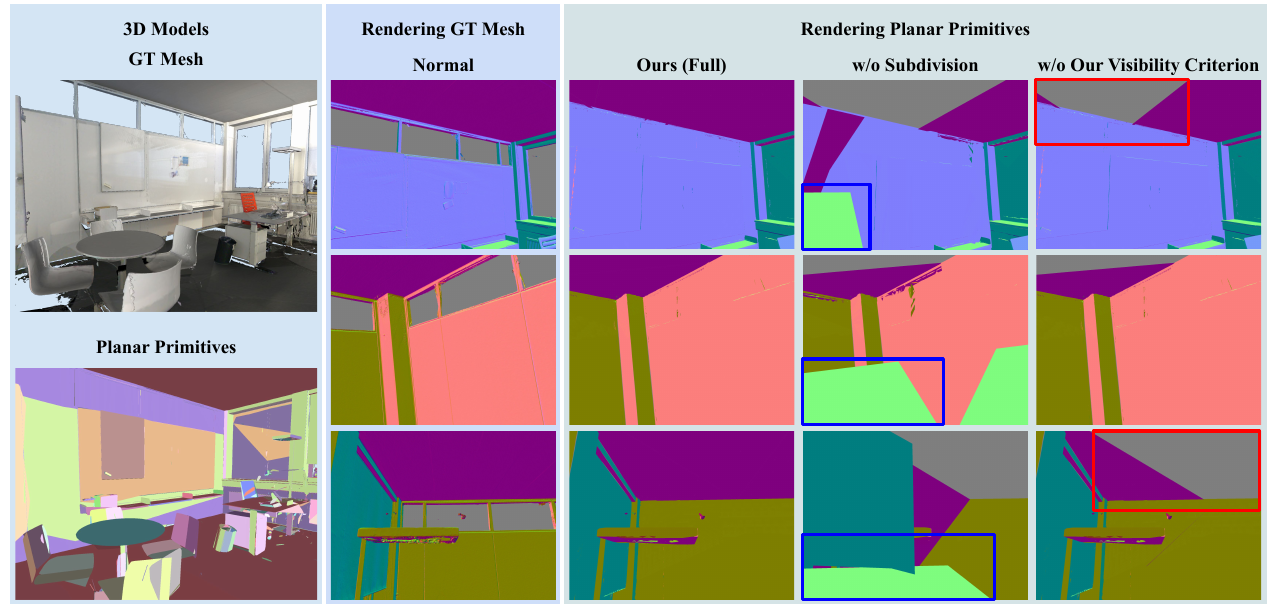}
   \vspace*{-6mm}
   \caption{
   \textbf{Results of forward rendering.} Planar primitives are extracted using GoCoPP~\cite{yu2022finding}. Each plane is assigned a specific color and is composed of multiple triangles. We utilize these large triangles to evaluate our forward rendering pipeline.
   }
   \label{fig:forward_rendering_result}
\end{figure}

Fig.~\ref{fig:forward_rendering_result} demonstrates the effectiveness of our forward rendering pipeline.  
The pipeline accurately reproduces depth and normal maps for both ground-truth meshes and our extracted planar primitives.  
The last two columns of Fig.~\ref{fig:forward_rendering_result} present an ablation study evaluating the impact of the proposed triangle subdivision and visibility operations.

\begin{figure}[!ht]
  \centering
\includegraphics[width=0.9\linewidth]{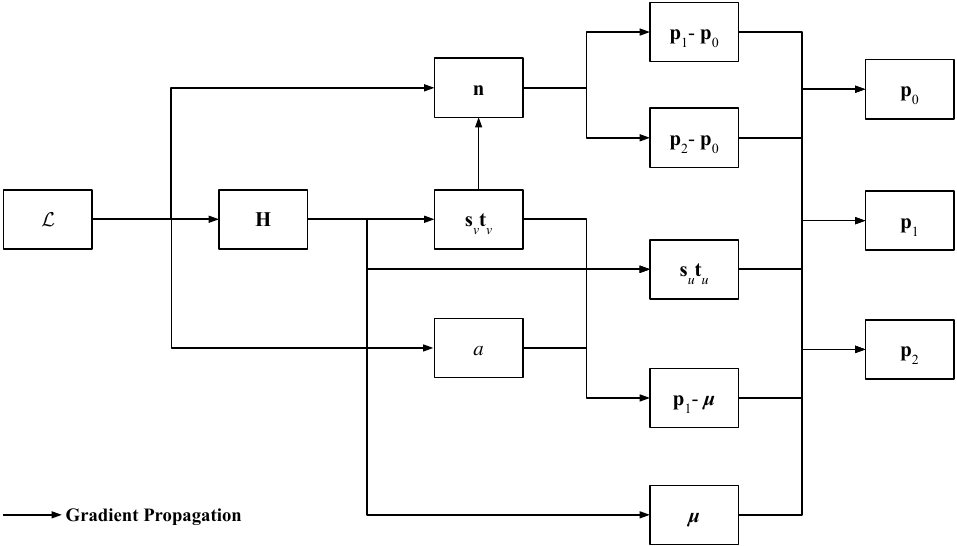}
   \vspace*{-2mm}
   \caption{
\textbf{An illustration of the backward computation graph.} This graph demonstrates the gradient propagation process from $\mathbf{H}$, $a$, and $\mathbf{n}$ to the gradients with respect to three vertices of the triangle $\mathbf{p}_0$, $\mathbf{p}_1$, and $\mathbf{p}_2$.}
   \label{fig:backward}
\end{figure}

\noindent \textbf{Backpropagation Process.}  
Thanks to the definition of the local tangent space in Eq.~2 in the main text, the coordinates of the three vertices of a triangle in the local tangent plane can be determined as
\begin{equation}
\mathbf{p}_0^{\prime} = (0, 1)^T,~~ \mathbf{p}_1^{\prime} = (a, 1)^T,~~
\mathbf{p}_2^{\prime} = (-1-a, -1)^T,
~~ \text{where}~~ a = \mathbf{t}_u \cdot (\mathbf{p}_1 - \bm{\mu}),
\label{eq:local_triangle_coor}
\end{equation}
The distance from the ray-triangle intersection $\hat{\mathbf{x}}$ to the three edges of the triangle $\mathbf{e}_1$, $\mathbf{e}_2$, and $\mathbf{e}_3$ in the local tangent plane can be determined solely based on $\hat{\mathbf{x}}$ and $a$ 
\begin{equation}
\begin{aligned}
\left\{
\begin{array}{l}
\mathrm{dist}(\hat{\mathbf{x}}, \mathbf{e}_0) = u+(1-a)v-1, \\[8pt]
\mathrm{dist}(\hat{\mathbf{x}}, \mathbf{e}_1) = -2u+(2a+1)v-1, \\[8pt]
\mathrm{dist}(\hat{\mathbf{x}}, \mathbf{e}_2) = u+(-2-a)v-1.
\end{array}
\right.
\quad \text{where} ~~ \hat{\mathbf{x}} = (u, v)^T
\end{aligned}.
\label{eq:local_triangle_frame}
\end{equation}
The definition of the local tangent space also simplifies the backpropagation process. For the gradients of $\alpha$, $\delta$, and $\sigma$, we follow the formulations of 2DGS~\cite{huang20242d} and 3DCS~\cite{held20243d}. According to the chain rule, the gradients with respect to the triangle vertices are illustrated in Fig.~\ref{fig:backward}.

\subsubsection{Triangle Pruning Mechanism} 
In Sec.~\ref{geo_training}, an efficient triangle pruning method is devised to attain clean and complete geometry. 
Entropy loss is introduced to the opacity $\alpha$ of each triangle in the last 5k iterations of the fine stage. Triangles with opacity $\alpha < 0.5$ and contribution $\gamma < 2.0$ are eliminated every 2k iterations. The contribution $\gamma_i$ of triangle $i$ during each 2k iterations can be computed as
\begin{equation}
   \gamma_i = \frac{1}{M}\sum_{k=1}^M\sum_{\mathbf{x}}  w\left(\hat{\mathbf{x}}_i\right) \prod_{j=1}^{i-1} \left(1 - w\left(\hat{\mathbf{x}}_j\right) \right),
\end{equation}
where $M$ represents the number of visible images of triangle $i$ during each 2k iterations. $\mathbf{x}$ denotes the pixel in image $k$ that the triangle $i$ covers. $\hat{\mathbf{x}}_i$ denotes the intersection point between the observation ray of pixel $\mathbf{x}$ and triangle $i$, and the calculation of $w\left(\hat{\mathbf{x}}_i\right)$ follows the details provided in Eq.~3 in the main text. Index $j$ refers to triangles sorted ahead of triangle $i$ from the perspective of image $k$.

\subsubsection{Orienting Triangle Normals}
\begin{figure}[!ht]
  \centering
\includegraphics[width=0.8\linewidth]{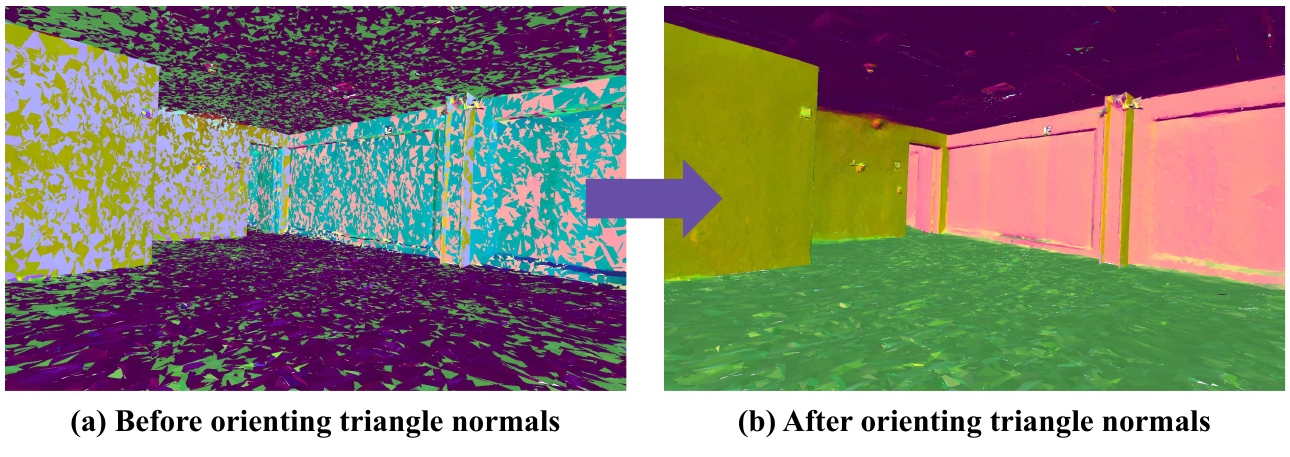}
   \vspace*{-2mm}
   \caption{
    \textbf{The effectiveness of orienting triangle normals.} The colors of the triangles are calculated based on their normals. 
   }
   \label{fig:oriented_normal}
\end{figure}

As discussed in Sec.~\ref{sec:octreegs}, we devise a post-processing algorithm to determine oriented normals of triangles, facilitating subsequent geometric reconstruction processes. 
The initial normal of each triangle is determined by Eq.~2 in the main text. For each triangle, we identify all images in which it is visible and flip its normal if it faces fewer than half of the observing views. 
As demonstrated in Fig.~\ref{fig:oriented_normal}, our approach effectively resolves normal orientation ambiguities, ensuring that surface normals consistently point toward the scene interior. 

\subsubsection{Configurations of LoD Plane Extraction} 
We perform LoD plane extraction using GoCoPP~\cite{yu2022finding}. The extraction process involves running GoCoPP ten times with coarse-to-fine parameter settings, where finer planar shapes are extracted from the remaining points after the coarser ones have been detected. The specific parameter configurations are provided in Table~\ref{tab:plane_detection}. The geometric tolerance parameter $\epsilon$ defines the maximum allowable distance between an inlier and its supporting plane, while $\sigma$ specifies the minimum inlier size required to retain a primitive. The normal consistency threshold $th_{\mathbf{n}}$ controls the maximum allowed deviation in normal direction.

\begin{table}[t!]
\centering
\renewcommand{\arraystretch}{1.15}
\setlength{\tabcolsep}{1pt}
   \caption{\textbf{Planar shape detection parameters}. Our plane extraction pipeline executes 10 iterations using a coarse-to-fine parameter strategy. $d$ is the bounding box diagonal of the scene.
   }
\label{tab:plane_detection}
\resizebox{0.8\linewidth}{!}{
\begin{tabular}{c|ccc|ccc|cccc}
\toprule
\multirow{4}{*}{Parameter} & \multicolumn{10}{c}{LoD 2} \\ 
\cmidrule(lr){2-7}
& \multicolumn{6}{c|}{LoD 1} & \multicolumn{4}{c}{} \\\cmidrule(lr){2-4} 
&\multicolumn{3}{c|}{LoD 0} & \multicolumn{3}{c|}{}  & \multicolumn{4}{c}{} \\
& Iter 0 & Iter 1 & Iter 2 & Iter 3 & Iter 4 & Iter 5 & Iter 6 & Iter 7 & Iter 8 & Iter 9 \\
\midrule
$\epsilon$ & 1.5\%$d$ & 0.2\%$d$ & 0.05\%$d$ & 0.05\%$d$ & 0.05\%$d$ & 0.05\%$d$ & 0.05\%$d$ & 0.05\%$d$ & 0.05\%$d$ & 0.05\%$d$ \\
$\sigma$ & 4000 & 500 & 200 & 100 & 80 & 30 & 20 & 10 & 5 & 4 \\
$th_{\mathbf{n}}$ & 0.85 & 0.85 & 0.85 & 0.8 & 0.8 & 0.8 & 0.75 & 0.75 & 0.7 & 0.5 \\
	
\bottomrule
\end{tabular}}
\end{table}

\subsection{More Experiments}
\label{sec:moreexp}
\subsubsection{Per-scene Rendering Results} 
We present the per-scene quality metrics (PSNR, SSIM, and LPIPS) utilized in our rendering evaluation in Sec.~\ref{exp:imple} for all considered methods, as shown from Table~\ref{tab:render_db_matrix_city} to Table~\ref{tab:render_fast_livo2}. Additionally, we provide the average number of rendered Gaussians across all test frames (\textit{\#Render}) and average storage size (\textit{\#Mem}).

\subsubsection{Per-scene Geometry Results} 
We list the per-scene quality metrics (Chamfer Distance and the total number of extracted planes) used in our geometry evaluation in Sec.~\ref{exp:imple} for all considered methods, as shown from Table~\ref{tab:geo_scannetpp} to Table~\ref{tab:geo_fast_livo2}. 

\subsubsection{More Results} 
\begin{figure}[!ht]
  \centering
\includegraphics[width=1\linewidth]{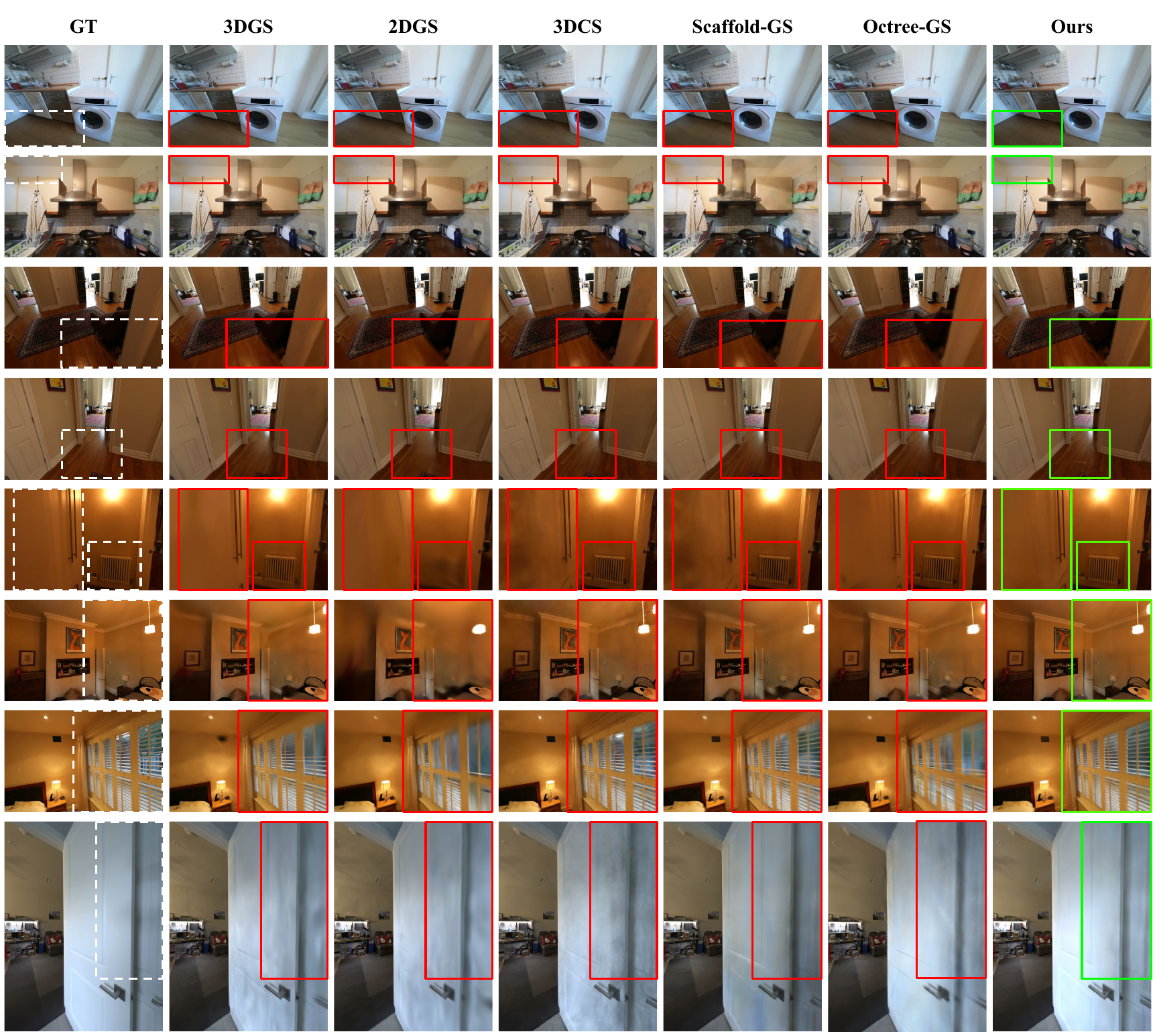}
   \vspace*{-6mm}
   \caption{
   \textbf{Qualitative rendering results.} We present the rendering results of additional scenes. Our method showcases superior rendering effects on surfaces such as walls, floors, and ceilings.}
   \label{fig:supp_rendering_compare}
\end{figure}

\begin{figure}[!ht]
  \centering
\includegraphics[width=1\linewidth]{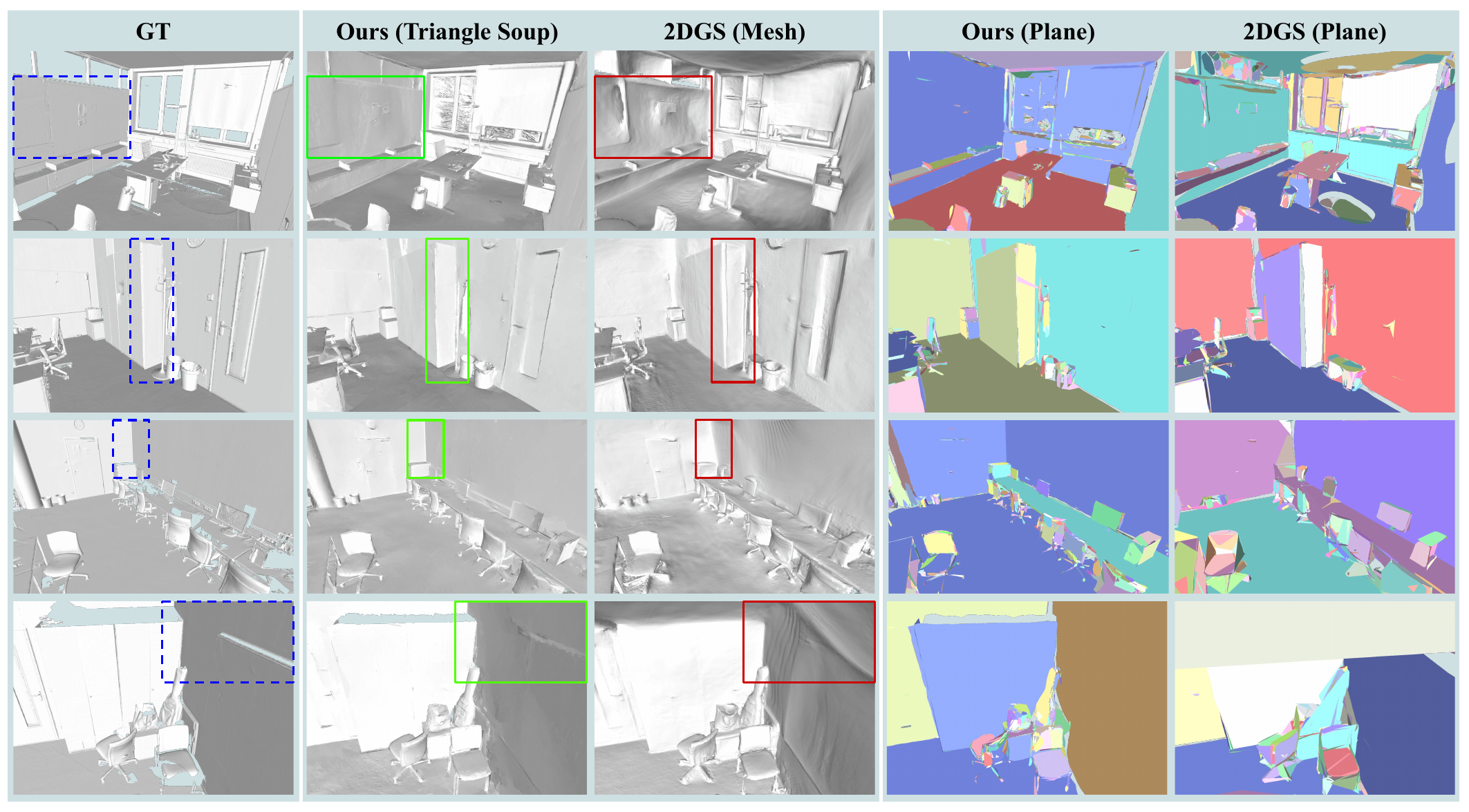}
   \vspace*{-4mm}
   \caption{
   \textbf{Qualitative geometric reconstruction comparisons.} We provide geometric reconstruction comparisons of four more scenes from ScanNet++~\cite{yeshwanth2023ScanNet++} dataset.}
   \label{fig:supp_geo_compare}
\end{figure}

\begin{figure}[!ht]
  \centering
\includegraphics[width=1\linewidth]{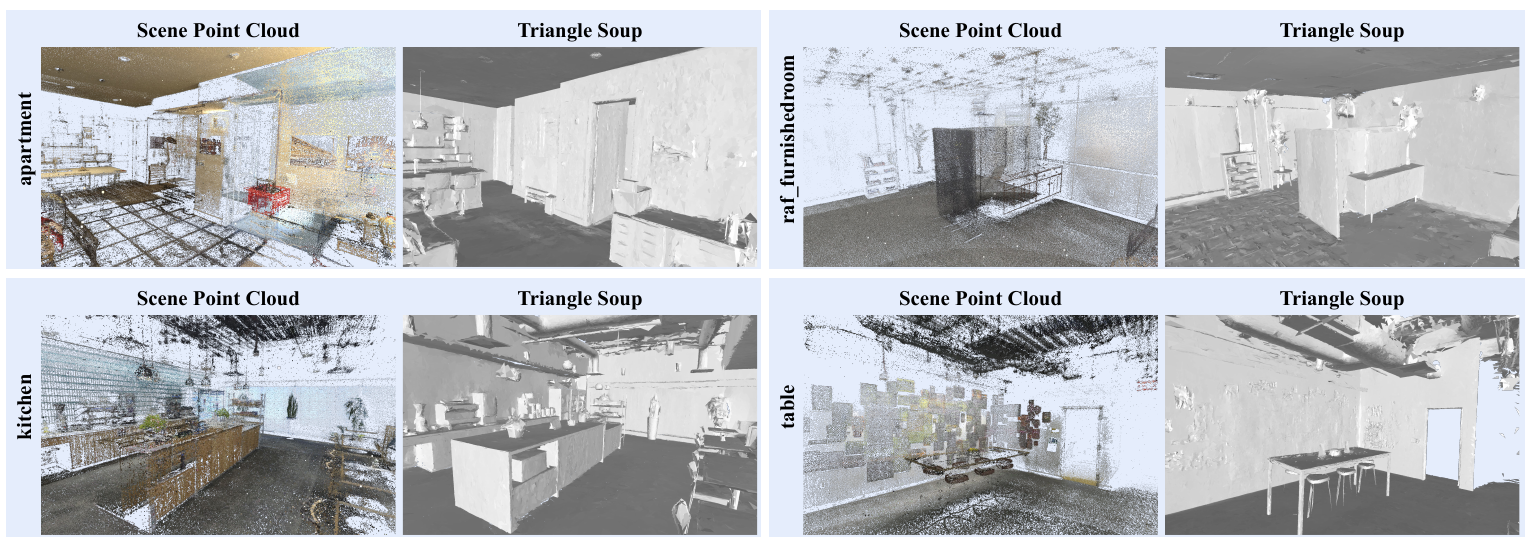}
   \vspace*{-4mm}
   \caption{
    \textbf{Qualitative geometric reconstruction results.} We present geometric reconstruction results of four more scenes from VR-NeRF~\cite{VRNeRF} dataset.
   }
   \label{fig:more_geo_vr_nerf}
\end{figure}

\begin{figure}[!ht]
  \centering
\includegraphics[width=1\linewidth]{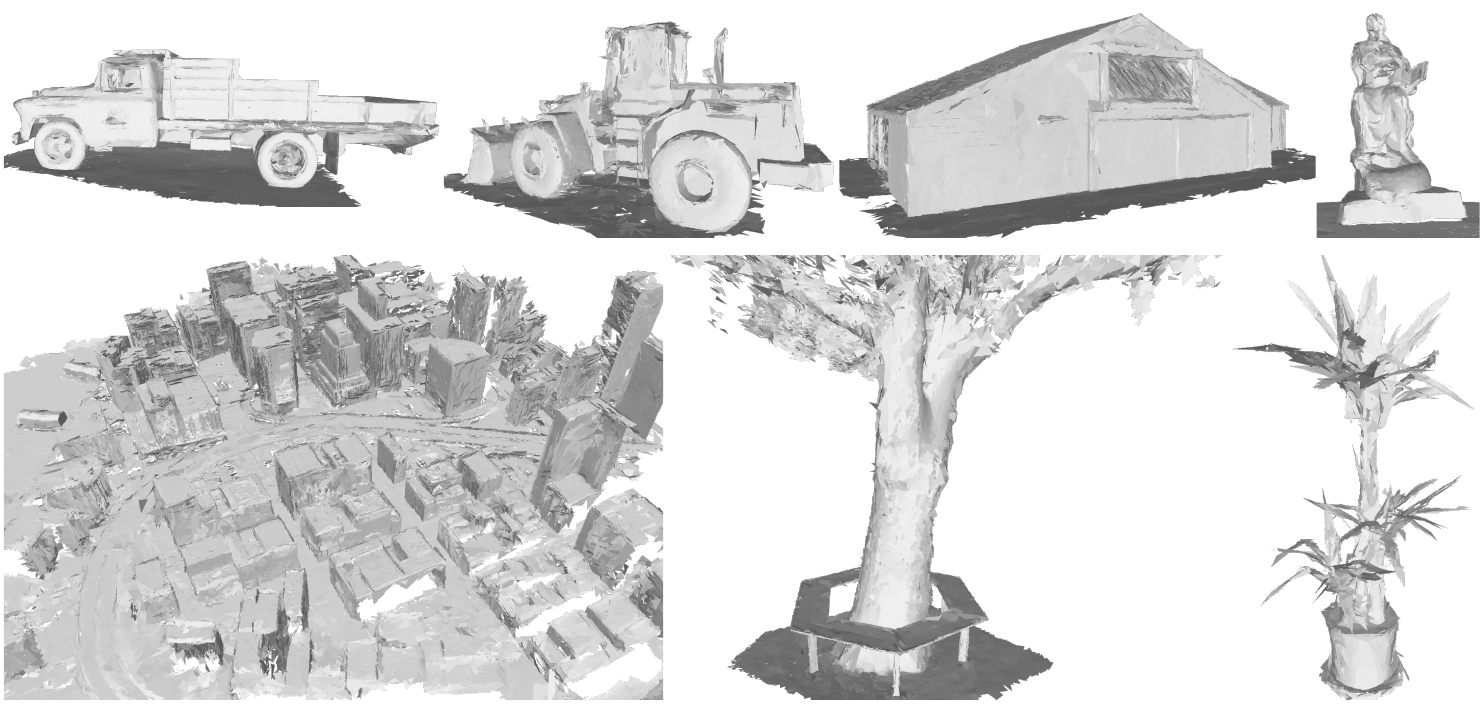}
   \vspace*{-4mm}
   \caption{We present additional geometric reconstruction results across diverse scenes from three datasets: Tanks\&Temples~\cite{knapitsch2017tanks}, MatrixCity~\cite{li2023matrixcity}, and MipNeRF‑360~\cite{barron2022mip}. The selected scenes encompass a comprehensive range of scales from large-scale outdoor environments to confined indoor objects, effectively demonstrating the robustness of our method.
   }
   \label{fig:more_geo_other}
   \vspace{-1em}
\end{figure}

In Fig.~\ref{fig:supp_rendering_compare}, we offer an additional visual comparison of our method with state-of-the-art 3DGS-based methods. Our renderings showcase superior quality. 
The additional qualitive geometric comparisons over ScanNet++~\cite{yeshwanth2023ScanNet++} dataset are shown in Fig.~\ref{fig:supp_geo_compare}. Fig.~\ref{fig:more_geo_vr_nerf} and~\ref{fig:more_geo_other} present more reconstruction results, demonstrating our method's robust applicability across diverse scenarios including both indoor and outdoor environments. Triangles serve as a powerful geometric representation, adept at capturing sharp object edges while still offering flexibility for representing curved surfaces.

\subsection{Limitations and Future Works}
\label{sec:limitation}
While our method exhibits strong reconstruction capabilities, several limitations remain. 
First, the current approach struggles with semi-transparent and transparent objects, as Neural Gaussians are unable to accurately capture their appearance. This results in an exaggerated and erroneous influence on the associated triangles.
Second, our framework primarily targets object surfaces and lacks dedicated treatment for the sky and distant background regions in outdoor scenes. This omission often leads to missing triangles in these areas, which in turn hinders appearance learning. In future work, we aim to introduce specialized models for background and sky representation—for example, by training additional Gaussians on a sky-sphere surrounding the scene.


\begin{table}[!ht]
\centering
\renewcommand{\arraystretch}{1.15}
\setlength{\tabcolsep}{1pt}
   \caption{\textbf{Rendering comparisons against baselines} over Deep Blending~\cite{hedman2018deep} scenes. Our method is compared with 3DGS~\cite{kerbl20233d}, 2DGS~\cite{huang20242d}, 3DCS~\cite{held20243d}, Scaffold-GS~\cite{lu2024scaffold}, and Octree-GS~\cite{ren2024octree}. 
   }
\label{tab:render_db_matrix_city}
\resizebox{0.85\linewidth}{!}{
\begin{tabular}{c|cccc|cccc}
\toprule
Scene & \multicolumn{4}{c|}{Dr Johnson} & \multicolumn{4}{c}{Playroom} \\
\begin{tabular}{c|c} Method & Metrics \end{tabular}  & PSNR$\uparrow$ & SSIM$\uparrow$ & LPIPS$\downarrow$ & \#Render/\#Mem & PSNR$\uparrow$ & SSIM$\uparrow$ & LPIPS$\downarrow$ & \#Render/\#Mem \\
\midrule
			
3DGS & 29.09 & 0.900 & 0.242 & 472K/ 818.9M & 29.83 & 0.905 & 0.241 & 324K/ 592.3M \\
           
2DGS & 28.74 & 0.897 & 0.257 & 232K/ 393.8M & 29.89 & 0.900 & 0.257 & 160K/ 276.7M \\
         
3DCS & 29.39 & 0.900 & 0.237 & \underline{171K}/ 330.3M & 29.87 & 0.901 & \underline{0.235} & \underline{129K}/ 243.9M \\

Scaffold-GS & \underline{29.73} & \textbf{0.910} & \underline{0.235} & 232K/ \textbf{145.0M} & \underline{30.74} & 0.907 & 0.242 & 182K/ 106.0M \\

Octree-GS & 29.71 & 0.908 & 0.237 & \textbf{113K}/ 156.2M & 30.64 & \underline{0.910} & 0.249 & \textbf{90K}/ \textbf{80.9M} \\

\hline

Ours & \textbf{29.75} & \underline{0.909} & \textbf{0.223} & 213K/ \underline{155.9M} & \textbf{30.86} & \textbf{0.913} & \textbf{0.229} & 172K/ \underline{103.7M} \\
\bottomrule
\end{tabular}}
\end{table}

\begin{table}[t!]
\centering
\renewcommand{\arraystretch}{1.15}
\setlength{\tabcolsep}{1pt}
   \caption{\textbf{Rendering comparisons against baselines} over Zip-NeRF~\cite{barron2023zipnerf} scenes. Our method is compared with 3DGS~\cite{kerbl20233d}, 2DGS~\cite{huang20242d}, 3DCS~\cite{held20243d}, Scaffold-GS~\cite{lu2024scaffold}, and Octree-GS~\cite{ren2024octree}. 
   }
\label{tab:render_zip_nerf}
\resizebox{1\linewidth}{!}{
\begin{tabular}{c|cccc|cccc|cccc|cccc}
\toprule
\multicolumn{1}{c|}{Scene} & \multicolumn{4}{c|}{alameda} & \multicolumn{4}{c|}{berlin} & \multicolumn{4}{c|}{london} & \multicolumn{4}{c}{nyc} \\
\begin{tabular}{c|c} Method & Metrics \end{tabular}  & PSNR$\uparrow$ & SSIM$\uparrow$ & LPIPS$\downarrow$ & \#Render/\#Mem & PSNR$\uparrow$ & SSIM$\uparrow$ & LPIPS$\downarrow$ & \#Render/\#Mem & PSNR$\uparrow$ & SSIM$\uparrow$ & LPIPS$\downarrow$ & \#Render/\#Mem & PSNR$\uparrow$ & SSIM$\uparrow$ & LPIPS$\downarrow$ & \#Render/\#Mem\\
\midrule		
2DGS & 20.45 & 0.669 & 0.482 & \textbf{100K}/ \textbf{94.6M} & 25.88 & 0.866 & 0.339 & \underline{126K}/ \underline{107.5M} & 24.10 & 0.759 & 0.442 & \textbf{80K}/ \textbf{74.4M} & 25.21 & 0.805 & 0.382 & \textbf{141K} / \underline{133.3M} \\   
3DGS & 22.12 & 0.717 & 0.420 & \underline{227K}/ 221.4M & 27.15 &  0.888 & 0.302 & 245K/ 213.3M & \underline{25.65} & 0.804 & 0.384 & \underline{188K}/ 181.1M & 26.63 & \underline{0.836} & 0.335 & 313K/ 302.3M \\  
3DCS & \underline{22.94} & \underline{0.738} & \underline{0.375} & 493K/ 526.2M & \textbf{27.93} & \textbf{0.897} & \textbf{0.272} & 394K/ 384.6M & 25.62 & 0.801 & 0.366 & 343K/ 356.5M & 26.34 & 0.832 & \underline{0.323} & 280K/ 301.5M \\
Scaffold-GS & 18.92 & 0.691 & 0.417 & 360K/ \underline{153.6M} & 26.21 & \underline{0.895} & \underline{0.285} & 422K/ 136.1M & 24.70 & 0.799 & 0.380 & 234K/ \underline{101.9M} & \underline{26.70} & \textbf{0.838} & 0.335 & 250K/ \textbf{126.5M} \\
Octree-GS & \textbf{23.02} & \textbf{0.752} & \textbf{0.368} & 342K/ 340.4M & 25.52 & 0.850 & 0.364 & \textbf{56K}/ \textbf{21.5M} & \textbf{26.04} & \textbf{0.821} & \textbf{0.351} & 238K/ 201.9M & 26.66 & 0.830 & 0.344 & \underline{168K}/ 187.2M \\
\hline
Ours & 22.76 & 0.731 & 0.388 & 371K/ 230.8M & \underline{27.61} & 0.894 & \underline{0.285} & 445K/ 210.5 &  25.64 & \underline{0.812} & \underline{0.353} & 526K/ 223.3M & \textbf{26.91} & \underline{0.836} & \textbf{0.322} & 308K/ 188.7M \\
\bottomrule
\end{tabular}}
\end{table}

\begin{table}[ht!]
\centering
\caption{\textbf{Rendering comparisons against baselines} over ScanNet++~\cite{yeshwanth2023ScanNet++} scenes. Our method is compared with 3DGS~\cite{kerbl20233d}, 2DGS~\cite{huang20242d}, 3DCS~\cite{held20243d}, Scaffold-GS~\cite{lu2024scaffold}, and Octree-GS~\cite{ren2024octree}. }
\label{tab:render_scannetpp}

\resizebox{1\linewidth}{!}{
\begin{tabular}{c|ccccccccccc}
\toprule
PSNR$\uparrow$ & 1ada7a0617 & 3e8bba0176 & 709ab5bffe & fe1733741f & 0a7cc12c0e & 8b5caf3398 & 0b031f3119 & 0cf2e9402d & a1d9da703c & 9859de300f \\ \midrule
			
3DGS & \underline{30.04} & \textbf{29.24} & \textbf{31.28} & 25.38 & \textbf{34.29} & \underline{31.33} & 27.92 & 32.48 & 26.59 & 35.74 \\
           
2DGS & 30.03 & 29.00 & 30.18 & 25.58 & 33.83 & 31.00 & \underline{28.16} & 31.06 & \underline{26.93} & 34.86 \\
         
3DCS & 29.99 & \textbf{29.24} & \underline{31.22} & \textbf{26.14} & 34.10 & 31.04 & \textbf{28.20} & \textbf{32.89} & \textbf{27.00} & \textbf{36.08} \\

Scaffold-GS & 27.87 & 28.08 & 25.80 & 25.31 & 33.50 & 29.44 & 27.05 & 29.50 & 26.30 & 33.14 \\

Octree-GS & \underline{30.04} & \textbf{29.24} & 30.50 & \underline{26.12} & \textbf{34.29} & \underline{31.33} & 27.92 & 32.48 & 26.59 & 34.19 \\

\hline

Ours & \textbf{30.34} & 29.10 & 30.92 & 25.30 & 34.12 & \textbf{31.46} & 28.03 & \underline{32.85} & \underline{26.93} & \underline{35.99} \\
\bottomrule
\end{tabular}}

\vspace{1.5em}

\resizebox{1\linewidth}{!}{
\begin{tabular}{c|ccccccccccc}
\toprule
SSIM$\uparrow$ & 1ada7a0617 & 3e8bba0176 & 709ab5bffe & fe1733741f & 0a7cc12c0e & 8b5caf3398 & 0b031f3119 & 0cf2e9402d & a1d9da703c & 9859de300f \\ \midrule
			
3DGS & \underline{0.937} & \textbf{0.932} & \textbf{0.928} & 0.858 & \textbf{0.964} & \textbf{0.945} & 0.908 & \underline{0.959} & \textbf{0.910} & 0.964 \\
           
2DGS & 0.935 & 0.929 & 0.919 & 0.861 & 0.960 & 0.944 & \underline{0.911} & 0.951 & \textbf{0.910} & 0.969 \\
         
3DCS & 0.935 & 0.931 & \underline{0.920} & \underline{0.868} & 0.961 & 0.943 & 0.909 & 0.952 & 0.909 & 0.966 \\

Scaffold-GS & 0.904 & 0.923 & 0.910 &  0.866 & 0.961 & 0.934 & 0.898 & 0.955 & 0.905 & \underline{0.970} \\

Octree-GS & \underline{0.937} & \textbf{0.932} & 0.916 & \textbf{0.874} & \textbf{0.964} & 0.943 & 0.908 & \underline{0.959} & 0.905 & 0.965 \\

\hline

Ours & \textbf{0.939} & 0.931 & 0.919 & 0.855 & 0.963 & \textbf{0.945} & \textbf{0.914} & \textbf{0.960} & 0.909 & \textbf{0.973} \\

\bottomrule
\end{tabular}}

\vspace{1.5em}

\resizebox{1\linewidth}{!}{
\begin{tabular}{c|ccccccccccc}
\toprule
LPIPS$\downarrow$ & 1ada7a0617 & 3e8bba0176 & 709ab5bffe & fe1733741f & 0a7cc12c0e & 8b5caf3398 & 0b031f3119 & 0cf2e9402d & a1d9da703c & 9859de300f \\ \midrule

3DGS & 0.176 & 0.149 & \underline{0.148} & 0.232 & \underline{0.126} & \underline{0.117} & \underline{0.202} & \underline{0.129} & 0.198 & 0.097 \\

2DGS & 0.189 & 0.161 & 0.166 & 0.231 & 0.136 & 0.126 & 0.207 & 0.155 & 0.210 & 0.100 \\

3DCS & 0.177 & \underline{0.145} & 0.162 & 0.207 & \underline{0.126} & 0.122 & 0.205 & 0.138 & \underline{0.196} & 0.095 \\

Scaffold-GS & \underline{0.167} & 0.152 & 0.167 & \underline{0.205} & 0.129 & 0.125 & 0.222 & 0.131 & 0.201 & \underline{0.089} \\

Octree-GS & 0.176 & 0.149 & 0.157 & \textbf{0.194} & \underline{0.126} & \underline{0.117} & \underline{0.202} & \underline{0.129} & 0.198 & 0.095 \\

\hline

Ours & \textbf{0.163} & \textbf{0.144} & \textbf{0.147} &  0.233 & \textbf{0.121} &  \textbf{0.112} & \textbf{0.189} & \textbf{0.111} &  \textbf{0.193}  & \textbf{0.079} \\
\bottomrule
\end{tabular}}

\vspace{1.5em}

\resizebox{1\linewidth}{!}{
\begin{tabular}{c|ccccccccccc}
\toprule
\#Render/\#Mem & 1ada7a0617 & 3e8bba0176 & 709ab5bffe & fe1733741f & 0a7cc12c0e & 8b5caf3398 & 0b031f3119 & 0cf2e9402d & a1d9da703c & 9859de300f \\ \midrule

3DGS & 231K/ 132M & 321K/ 223.9M & 313K/ 212.5M & 498K/ 376.9M & 290K/ 276.2M & 281K/ 156.8M & 267K/ 211.3M & 378K/ 241.4M & 309K/ 204.7M & 378K/ 255M \\

2DGS & \underline{124K}/ 69.9M & 181K/ 123.6M & 167K/ 115.1M & 251K/ 191.3M & 150K/ 140.8M & 149K/ 82.9M & \underline{145K}/ 111.7M & \underline{178K}/ 114M & \textbf{174K}/ 112.4M & 178K/ 125.8M \\

3DCS & \textbf{102K}/ \underline{67.6M} & \textbf{133K}/ 97.5M & \textbf{70K}/ \underline{48.9M} & \textbf{162K}/ \underline{137.4M} & \textbf{68K}/ \underline{66.3M} & \textbf{83K}/ \textbf{53M} & \textbf{109K}/ \underline{95.4M} & \textbf{86K}/ \textbf{62.6M} & \underline{185K}/ 133.3M & \textbf{55K}/ \underline{38.1M} \\

Scaffold-GS & 184K/ 103M & \underline{169K}/ \underline{81.9M} & 199K/ 49.9M & 386K/ 185.2M & \underline{139K}/ \textbf{65.2M} & \underline{140K}/ \underline{71.1M} & 152K/ \textbf{88.4M} & 192K/ \underline{64.5M} & 207K/ \underline{110.3M} & 180K/ 56.2M \\

Octree-GS & 231K/ 132M & 321K/ 223.9M & \underline{135K}/ \textbf{27.4M} & 385K/ 210.7M & 290K/ 276.2M & 281K/ 156.8M & 267K/ 211.3M & 378K/ 241.4M & 309K/ 204.7M & \underline{84K}/ \textbf{21.1M} \\

\hline

Ours &  257K/ \textbf{65.2M} & 310K/ \textbf{72.6M} & 346K/ 85.3M &  \underline{174K}/ \textbf{69.3M} & 316K/ 107.4M & 407K/ 124.4M & 343K/ 131.9M & 487K/ 159.5M &  216K/ \textbf{78.2M} & 500K/ 126.1M \\
\bottomrule
\end{tabular}} 
\end{table}

\begin{table}[ht!]
\centering
\caption{\textbf{Rendering comparisons against baselines} over VR-NeRF~\cite{VRNeRF} scenes. Our method is compared with 3DGS~\cite{kerbl20233d}, 2DGS~\cite{huang20242d}, 3DCS~\cite{held20243d}, Scaffold-GS~\cite{lu2024scaffold}, and Octree-GS~\cite{ren2024octree}. }
\label{tab:render_vr_nerf}

\resizebox{1\linewidth}{!}{
\begin{tabular}{c|ccccccccccc}
\toprule
PSNR$\uparrow$ & apartment & kitchen & office\_view1 & office\_view2 & office1b & raf\_emptyroom & raf\_furnishedroom & riverview & seating\_area & table & workshop \\ \midrule
			
2DGS & 29.92 & 30.65 & 28.45 & 30.14 & 27.56 & 30.20 & 30.29 & 23.27 & 32.23 & 30.56 & 29.09 \\
           
3DGS & 30.84 & 31.59 & 29.16 & 30.94 & 28.18 & 30.60 & 31.21 & 24.51 & 33.26 & 31.44 &  \underline{30.42} \\
         
3DCS & \textbf{31.42} & \textbf{32.02} & \underline{30.52} & \underline{31.72} & 28.62 & \underline{31.72} & \underline{31.68} & 24.28 & 33.15 & 30.98 & \textbf{30.95} \\

Scaffold-GS & 30.99 & 31.59 & 29.94 & 31.44 & 27.39 & 30.57 & \textbf{31.71} & \underline{26.05} & \underline{33.72} & 30.92 & 29.76 \\

Octree-GS & 29.71 & 30.64 & \textbf{30.64} & 28.34 & \textbf{31.04} & \textbf{32.02} & 28.68 & \textbf{26.10} & 31.47 & \textbf{34.51} & 28.18 \\

\hline

Ours & \underline{31.12} & \underline{31.64} & 30.46 & \textbf{31.79} & \underline{28.73} & 31.62 & 31.46 & 25.93 & \textbf{34.11} & \underline{32.30} & 30.05 \\
\bottomrule
\end{tabular}}

\vspace{1.5em}

\resizebox{1\linewidth}{!}{
\begin{tabular}{c|ccccccccccc}
\toprule
SSIM$\uparrow$ & apartment & kitchen & office\_view1 & office\_view2 & office1b & raf\_emptyroom & raf\_furnishedroom & riverview & seating\_area & table & workshop \\ \midrule
			
2DGS & 0.910 & 0.919 & 0.917 & 0.907 & 0.877 & 0.887 & 0.872 & 0.855 & 0.943 & 0.912 & 0.893 \\
           
3DGS & \underline{0.926} & \underline{0.935} & 0.929 & \underline{0.935} & 0.898 & 0.915 & 0.905 & 0.869 & \underline{0.954} & \underline{0.923} & \underline{0.913} \\
         
3DCS & 0.925 & 0.932 & 0.926 & 0.922 & 0.884 & 0.896 & 0.878 & 0.849 & 0.948 & 0.912 & \textbf{0.919} \\

Scaffold-GS & \textbf{0.934} & \textbf{0.939} & \textbf{0.937} & \textbf{0.948} & \underline{0.909} & \underline{0.917} & \textbf{0.920} & \textbf{0.883} & \textbf{0.957} & 0.921 & 0.911 \\

Octree-GS & 0.908 & 0.910 & 0.901 & 0.891 & \textbf{0.937} & \textbf{0.940} & \underline{0.912} & \underline{0.877} & 0.902 & \textbf{0.957} & 0.899 \\

\hline

Ours & 0.922 & 0.930 &  \underline{0.930} & 0.932 & 0.895 & 0.907 & 0.899 & \underline{0.877} & 0.951 & 0.922 & 0.908 \\
\bottomrule
\end{tabular}}

\vspace{1.5em}

\resizebox{1\linewidth}{!}{
\begin{tabular}{c|ccccccccccc}
\toprule
LPIPS$\downarrow$ & apartment & kitchen & office\_view1 & office\_view2 & office1b & raf\_emptyroom & raf\_furnishedroom & riverview & seating\_area & table & workshop \\ \midrule    
2DGS & 0.252 & 0.225 & 0.273 & 0.252 & 0.243 & 0.263 & 0.279 & 0.248 & 0.180 & 0.229 & 0.267 \\
3DGS & \underline{0.193} & \underline{0.171} & \underline{0.216} & \underline{0.177} &0.192 & 0.199 & 0.208 & 0.201 & \underline{0.136} &  0.181 & 0.207 \\
3DCS & 0.212 & 0.187 & 0.251 & 0.200 & 0.217 & 0.218 & 0.240 & 0.257 & 0.150 & 0.195 & \textbf{0.197}\\
Scaffold-GS & \textbf{0.154} & \textbf{0.149} & \textbf{0.175} & \textbf{0.131} & \textbf{0.154} & \underline{0.182} & \underline{0.166} & \textbf{0.177} & \textbf{0.116} & \underline{0.168} & \underline{0.200} \\
Octree-GS & 0.237 & 0.249 & 0.256 & 0.202 & \underline{0.186} & \textbf{0.158} & \textbf{0.151} & \underline{0.197} & 0.207 & \textbf{0.116} & 0.239 \\
\hline
Ours & 0.220 & 0.193 & 0.230 & 0.189 & 0.200 & 0.205 & 0.210 & 0.198 & 0.149 & 0.186 & 0.228 \\
\bottomrule
\end{tabular}}

\vspace{1.5em}

\resizebox{1\linewidth}{!}{
\begin{tabular}{c|ccccccccccc}
\toprule
\#Render/\#Mem & apartment & kitchen & office\_view1 & office\_view2 & office1b & raf\_emptyroom & raf\_furnishedroom & riverview & seating\_area & table & workshop \\ \midrule
2DGS & \underline{145K}/ \underline{181.3M} & \underline{162K}/ \underline{183.0M} & \textbf{62K}/ \underline{76.4M} & \textbf{129K}/ \underline{131.2M} & \textbf{142K}/ \textbf{124.6M} & \underline{110K}/ \textbf{112.5M} & \textbf{117K}/ \textbf{127.3M} & \underline{235K}/ 299.4M & \textbf{170K}/ \textbf{214.3M} & \textbf{162K}/ \textbf{199.1M} & \textbf{152K}/ \underline{172.4M} \\
3DGS & 535K/ 724.6M & 522K/ 621.5M & 227K/ 290.7M & 501K/ 552.7M & 422K/ 382.1M & 442K/ 474.3M & 465K/ 535.3M & 590K/ 784.5M & 432K/ 561.7M & 417K/ 533.2M & 478K/ 565.4M \\
3DCS & 351K/ 481.5M & 343K/ 450.7M & 126K/ 167.6M & 327K/ 380.9M & 235K/ 233.0M & 181K/ 208.1M & \underline{198K}/ 232.2M & \textbf{141K}/ \underline{185.6M} & \underline{251K}/ 341.9M & \underline{236K}/ 327.6M & 519K/ 662.3M \\
Scaffold-GS & 1139K/ 872.9M & 766K/ 686.5M & 614K/ 323.2M & 1240K/ 684.4M & 1076K/ 358.0M & 711K/ 254.0M & 1127K/ 582.2M & 550K/ 327.9M & 717K/ 422.1M & 624K/ 332.0M & 781K/ 642.5M \\
Octree-GS & \textbf{113K}/ \textbf{156.2M} & \textbf{90K}/ \textbf{80.9M} & \underline{82K}/ \textbf{58.9M} & \underline{277K}/ \textbf{111.6M} & \underline{198K}/ 495.2M & \textbf{64K}/ 786.1M & 691K/ 796.3M & 264K/ \textbf{135.1M} & 293K/ 750.7M & 483K/ 747.2M & \underline{407K}/ \textbf{142.1M} \\
\hline
Ours & 360K/ 212.6M & 371K/ 205.0M & 232K/ 119.6M & 360K/ 154.3M & 363K/ \underline{166.0M} & 441K/ \underline{170.1M} & 584K/ \underline{225.0M} & 458K/ 288.6M & 574K/ \underline{217.5M} & 501K/ \underline{233.1M} & 464K/ 188.1M \\
\bottomrule
\end{tabular}}
\end{table}

\begin{table}[t!]
\centering
\renewcommand{\arraystretch}{1.15}
\setlength{\tabcolsep}{1pt}
   \caption{\textbf{Rendering comparisons against baselines} over FAST-LIVO2~\cite{zheng2024fast} scenes. Our method is compared with 3DGS~\cite{kerbl20233d}, 2DGS~\cite{huang20242d}, 3DCS~\cite{held20243d}, Scaffold-GS~\cite{lu2024scaffold}, and Octree-GS~\cite{ren2024octree}. 
   }
\label{tab:render_fast_livo2}
\resizebox{1\linewidth}{!}{
\begin{tabular}{c|cccc|cccc|cccc|cccc}
\toprule
\multicolumn{1}{c|}{Scene} & \multicolumn{4}{c|}{CBD\_Building\_01} & \multicolumn{4}{c|}{HKU\_Landmark} & \multicolumn{4}{c|}{Retail\_Street} & \multicolumn{4}{c}{SYSU\_01} \\
\begin{tabular}{c|c} Method & Metrics \end{tabular}  & PSNR$\uparrow$ & SSIM$\uparrow$ & LPIPS$\downarrow$ & \#Render/\#Mem & PSNR$\uparrow$ & SSIM$\uparrow$ & LPIPS$\downarrow$ & \#Render/\#Mem & PSNR$\uparrow$ & SSIM$\uparrow$ & LPIPS$\downarrow$ & \#Render/\#Mem & PSNR$\uparrow$ & SSIM$\uparrow$ & LPIPS$\downarrow$ & \#Render/\#Mem\\
\midrule			
2DGS & 28.25 & 0.896 & 0.150 & 511K/ 694.0M & 28.78 & 0.780 & 0.302 & 395K/ 376.1M & 28.54 & 0.895 & 0.126 & 674K/ 1239.9M & 27.94 & 0.837 & 0.225 & \underline{210K}/ 333.8M \\        
3DGS & 29.20 & 0.907 & 0.116 & 558K/ 721.8M & 29.92 & 0.803 & 0.260 & 340K/ 405.0M & \textbf{30.68} & \textbf{0.920} & 0.090 & 732K/ 1045.8M & 30.04 & 0.876 & 0.156 & 448K/ 868.2M \\   
3DCS & \textbf{29.36} & 0.906 & 0.113 & \textbf{235K}/ 199.5M & 29.57 & 0.802 & 0.248 & \textbf{254K}/ 158.8M & \underline{30.23} & 0.914 & \textbf{0.084} & \textbf{198K}/ 274.9M & 30.24 & 0.877 & 0.143 & \textbf{182K}/ 271.7M \\
Scaffold-GS & 24.86 & 0.846 & 0.138 & \underline{325K}/ 197.9M & 28.63 & 0.759 & 0.254 & \underline{328K}/ \textbf{141.3M} & 28.04 & 0.869 & 0.107 & 427K/ 295.2M & 27.06 & 0.834 & 0.170 & 518K/ 266.5M \\
Octree-GS & 29.25 & \underline{0.909} & \textbf{0.105} & 561K/ \underline{165.4M} & \textbf{30.08} & \textbf{0.817} & \textbf{0.229} & 507K/ 191.5M & 29.63 & 0.916 & 0.092 & \underline{324K}/ \textbf{170.0M} & \textbf{30.70} & \textbf{0.886} & \underline{0.129} & 436K/ \textbf{172.5M} \\
\hline
Ours & \underline{29.26} & \textbf{0.910} & \underline{0.106} & 409K/ \textbf{161.3M} & \underline{29.97} & \underline{0.816} & \textbf{0.229} & 449K/ \underline{143.9M} & 29.66 & \textbf{0.920} & \underline{0.086} & 508K/ \underline{269.1M} & \underline{30.33} & \textbf{0.886} & \textbf{0.128} & 622K/ \underline{234.1M} \\
\bottomrule
\end{tabular}}
\end{table}

\begin{table}[ht!]
\centering
\caption{\textbf{Geometry comparisons against baselines} over ScanNet++~\cite{yeshwanth2023ScanNet++} scenes. Our method is compared with MVS~\cite{schoenberger2016mvs}, 2DGS~\cite{huang20242d}, and GFSGS~\cite{jiang2024geometry}. }
\label{tab:geo_scannetpp}
\resizebox{\linewidth}{!}{
\begin{tabular}{c|cccccccccc}
\toprule
Ch\_L2(cm) $\downarrow$ & 1ada7a0617 & 3e8bba0176 & 709ab5bffe & fe1733741f & 0a7cc12c0e & 8b5caf3398 & 0b031f3119 & 0cf2e9402d & a1d9da703c & 9859de300f \\ \midrule
MVS & 20.52 & 19.39 & 13.79 & 24.71 & 29.28 & 13.61 & 16.10 & 14.08 & 9.95 & 8.57 \\
2DGS & \underline{15.64} & \underline{18.41} & \underline{11.13} & \underline{10.62} & \underline{7.22} & \underline{6.90} & \underline{15.72} & \underline{12.17} & \underline{8.79} & \textbf{5.87} \\
GFSGS & 24.56 & 27.99 & 14.34 & 12.92 & 17.12 & 12.39 & 16.37 & 19.40 & 14.12 & 13.68 \\
Ours & \textbf{7.26} & \textbf{11.72} & \textbf{7.44} & \textbf{9.87} & \textbf{5.37} & \textbf{6.09} & \textbf{7.97} & \textbf{7.29} & \textbf{8.19} & \underline{7.19} \\
\bottomrule
\end{tabular}
}

\vspace{1.5em}

\resizebox{\linewidth}{!}{
\begin{tabular}{c|cccccccccc}
\toprule
\# Planar Primitives $\downarrow$ & 1ada7a0617 & 3e8bba0176 & 709ab5bffe & fe1733741f & 0a7cc12c0e & 8b5caf3398 & 0b031f3119 & 0cf2e9402d & a1d9da703c & 9859de300f \\ \midrule
MVS & 9851 & 10735 & 8013 & \underline{8177} & 8319 & 10506 & 8978 & 9760 & 10268 & 11980 \\
2DGS & \textbf{619} & \textbf{1358} & \underline{5066} & 13040 & \textbf{807} & \underline{2822} & \textbf{1800} & \underline{4268} & \underline{9193} & \textbf{703} \\
GFSGS & 5026 & 5045 & 6990 & 44677 & 5230 & 3926 & 10602 & 4415 & 14932 & 5656 \\
Ours & \underline{1454} & \underline{2470} & \textbf{2324} & \textbf{2902} & \underline{1773} & \textbf{1039} & \underline{2033} & \textbf{1969} & \textbf{2314} & \underline{1383} \\
\bottomrule
\end{tabular}
}
\end{table}

\begin{table}[ht!]
\vspace{-1.5em}
\centering
\caption{\textbf{Geometry comparisons against baselines} over FAST-LIVO2~\cite{zheng2024fast} scenes. Our method is compared with MVS~\cite{schoenberger2016mvs}, 2DGS~\cite{huang20242d}, and GFSGS~\cite{jiang2024geometry}. }
\label{tab:geo_fast_livo2}
\resizebox{1\linewidth}{!}{
\begin{tabular}{c|cc|cc|cc|cc}
\toprule
\multicolumn{1}{c|}{Scene} & \multicolumn{2}{c|}{CBD\_Building\_01} & \multicolumn{2}{c|}{HKU\_Landmark} & \multicolumn{2}{c|}{Retail\_Street} & \multicolumn{2}{c}{SYSU\_01} \\
\multicolumn{1}{c|}{Method \& Metrics} & Ch\_L2(cm) $\downarrow$ & \# Planar Primitives $\downarrow$ & Ch\_L2(cm) $\downarrow$ & \# Planar Primitives $\downarrow$ & Ch\_L2(cm) $\downarrow$ & \# Planar Primitives $\downarrow$ & Ch\_L2(cm) $\downarrow$ & \# Planar Primitives $\downarrow$ \\
\midrule
MVS & 40.75 & 5435 & 31.47 & 4945 & 30.31 & 4548 & \textbf{24.19} & 5810 \\
2DGS & 42.08 & \underline{2897} & 42.09 & \textbf{1183} & 36.54 & \underline{3446} & 45.99 & \textbf{3188} \\
GFSGS & \underline{29.01} & \textbf{2065} & \textbf{26.82} & 4602 & \textbf{24.12} & 3821 & \underline{25.54} & \underline{3264} \\
Ours & \textbf{19.72} & 3783 & \underline{30.69} & \underline{4415} & \underline{28.20} & \textbf{2118} & 25.62 & 4233 \\
\bottomrule
\end{tabular}}
\vspace{-2em}
\end{table}

\clearpage